%% file: main.tex
\documentclass{article} % For LaTeX2e
\usepackage{iclr2020_conference,times}

\input{math_commands.tex}

\usepackage{hyperref}
\usepackage{url}
\usepackage{booktabs}  % professional-quality tables

\usepackage[T1]{fontenc}
\usepackage[english]{babel}
\usepackage[pdftex]{graphicx}
\usepackage{subfigure}
\usepackage{wrapfig}
\usepackage{caption}
\usepackage{pdflscape}
\usepackage{afterpage}
\usepackage{algorithm}
\usepackage{algorithmicx}
\usepackage{algpseudocode}
\usepackage{nicefrac}  % compact symbols for 1/2, etc.
\usepackage{color}  % to highlight notes

\algnewcommand\algorithmicinput{\textbf{Input:}}
\algnewcommand\algorithmicoutput{\textbf{Output:}}
\algnewcommand\Input{\item[\algorithmicinput]}%
\algnewcommand\Output{\item[\algorithmicoutput]}%

\newcommand{\lab}{\gamma_{l}}
\newcommand{\pol}{\gamma_{p}}
\newcommand{\klab}{k_{l}}

\title{Deep Semi-Supervised Anomaly Detection}

\author{%
Lukas Ruff\textsuperscript{\textnormal 1} \qquad Robert A.~Vandermeulen\textsuperscript{\textnormal 1}\thanks{ Majority of the work was done while RV was at TU Kaiserslautern, Germany.} \qquad Nico G{\"o}rnitz\textsuperscript{\textnormal{ 1\,2}} \\
  \textbf{Alexander Binder}\textsuperscript{3} \qquad \textbf{Emmanuel M{\"u}ller}\textsuperscript{4} \\
  \textbf{Klaus-Robert M{\"u}ller}\textsuperscript{1\,5\,6} \qquad \textbf{Marius Kloft}\textsuperscript{7}\thanks{ Part of the work was done while MK was a sabbatical visitor of the DASH Center at the University of Southern California, United States.} \\
  \textsuperscript{1}Technical University of Berlin, Germany \\
  \textsuperscript{2}123ai.de, Berlin, Germany \\
  \textsuperscript{3}Singapore University of Technology \& Design, Singapore \\
  \textsuperscript{4}Bonn-Aachen International Center for Information Technology, Germany \\
  \textsuperscript{5}Korea University, Seoul, Republic of Korea \\
  \textsuperscript{6}Max Planck Institute for Informatics, Saarbr{\"u}cken, Germany \\
  \textsuperscript{7}Technical University of Kaiserslautern, Germany \\
  \texttt{\{lukas.ruff, vandermeulen, nico.goernitz\}@tu-berlin.de} \\
  \texttt{alexander\_binder@sutd.edu.sg} \qquad \texttt{mueller@bit.uni-bonn.de} \\
  \texttt{klaus-robert.mueller@tu-berlin.de} \qquad \texttt{kloft@cs.uni-kl.de}%
}

\iclrfinalcopy % Uncomment for camera-ready version, but NOT for submission.
\begin{document}

\maketitle

\begin{abstract}
Deep approaches to anomaly detection have recently shown promising results over shallow methods on large and complex datasets.
Typically anomaly detection is treated as an unsupervised learning problem.
In practice however, one may have---in addition to a large set of unlabeled samples---access to a small pool of labeled samples, e.g.~a subset verified by some domain expert as being normal or anomalous.
Semi-supervised approaches to anomaly detection aim to utilize such labeled samples, but most proposed methods are limited to merely including labeled normal samples.
Only a few methods take advantage of labeled anomalies, with existing deep approaches being domain-specific.
In this work we present \emph{Deep SAD}, an end-to-end deep methodology for general semi-supervised anomaly detection.
We further introduce an information-theoretic framework for deep anomaly detection based on the idea that the entropy of the latent distribution for normal data should be lower than the entropy of the anomalous distribution, which can serve as a theoretical interpretation for our method.
In extensive experiments on MNIST, Fashion-MNIST, and CIFAR-10, along with other anomaly detection benchmark datasets, we demonstrate that our method is on par or outperforms shallow, hybrid, and deep competitors, yielding appreciable performance improvements even when provided with only little labeled data.
\end{abstract}

\section{Introduction}
\label{sec:introduction}

%%%%%%%%%%%%%%%%%%%%%%%%%%%%%%%%%%%%%%%%%%%%%%%%%%%%%%%%%%%%%%%%%%%%%%%%%%%%%%%%
\begin{figure}[t]
\begin{center}
\vspace{-0.5em}
\includegraphics[width=\linewidth]{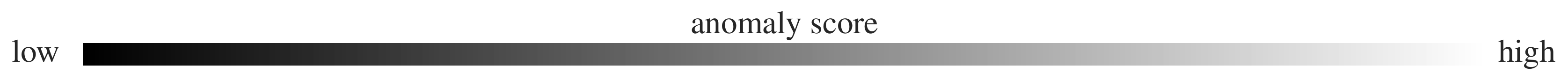}
%\vspace{-0.5em}
\subfigure[Training data]{\label{fig:toy_train}\includegraphics[width=0.329\linewidth]{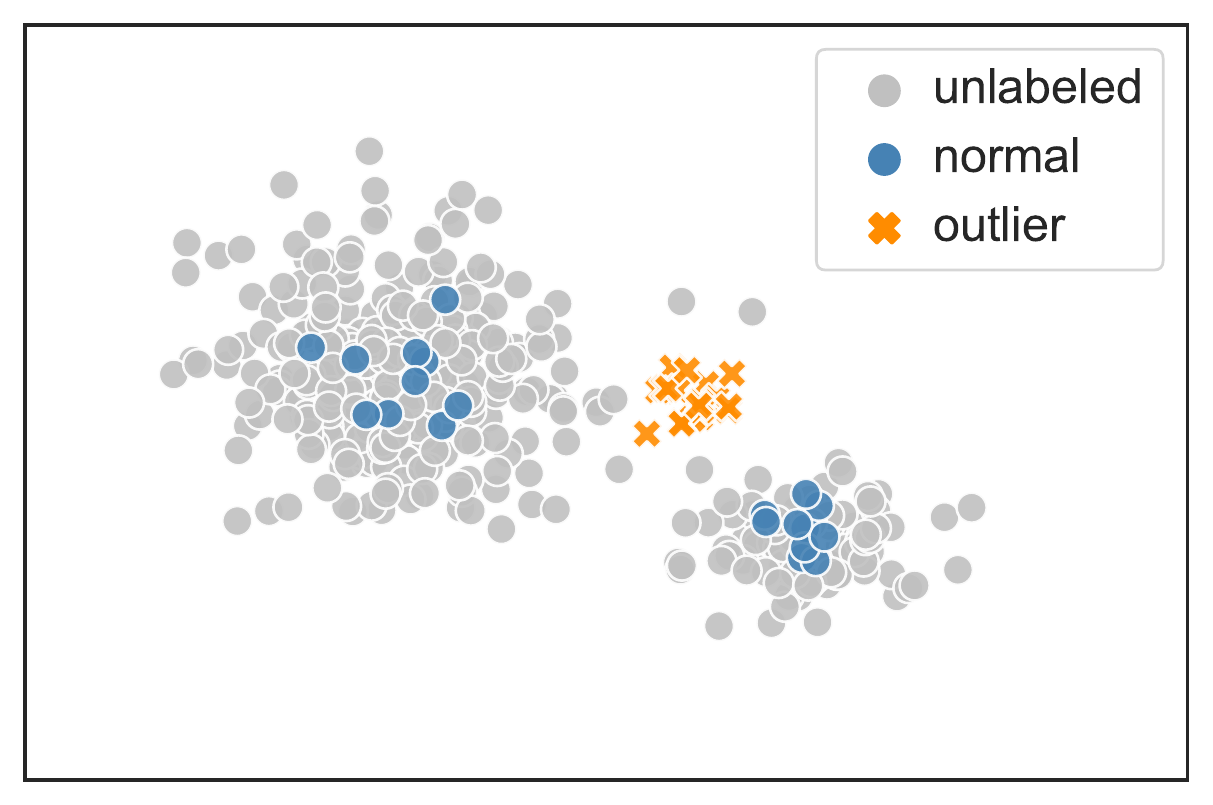}}
\subfigure[Unsupervised AD (OC-SVM)]{\label{fig:toy_unsupervised}\includegraphics[width=0.329\linewidth]{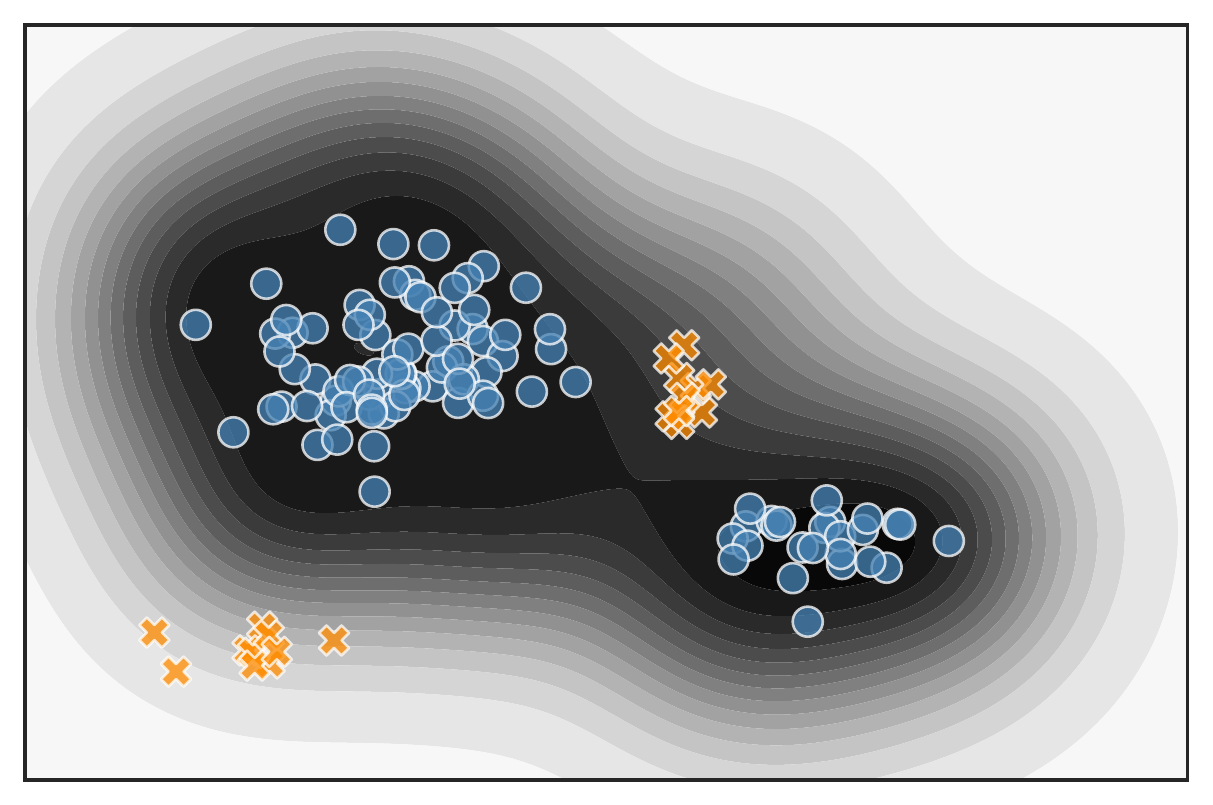}}
\subfigure[Supervised classifier (SVM)]{\label{fig:toy_supervised}\includegraphics[width=0.329\linewidth]{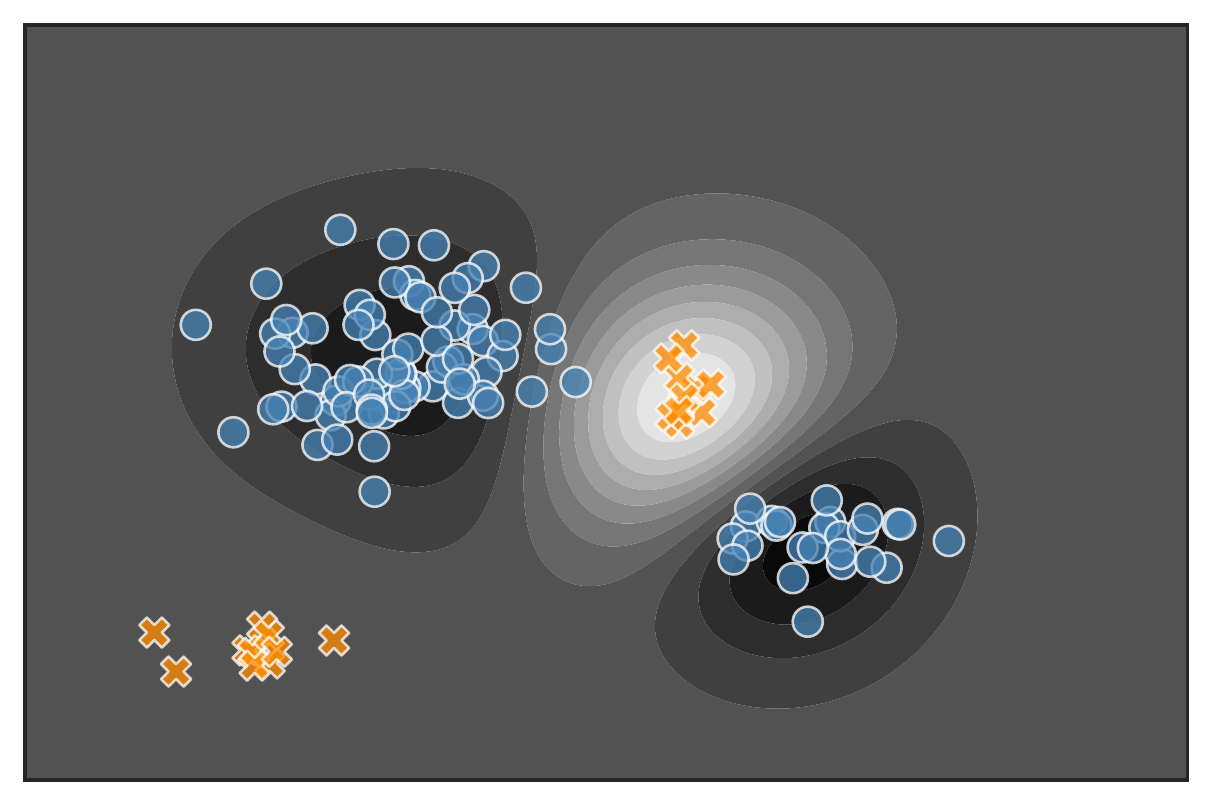}}
\subfigure[Semi-supervised classifier]{\label{fig:toy_semisupervised_class}\includegraphics[width=0.329\linewidth]{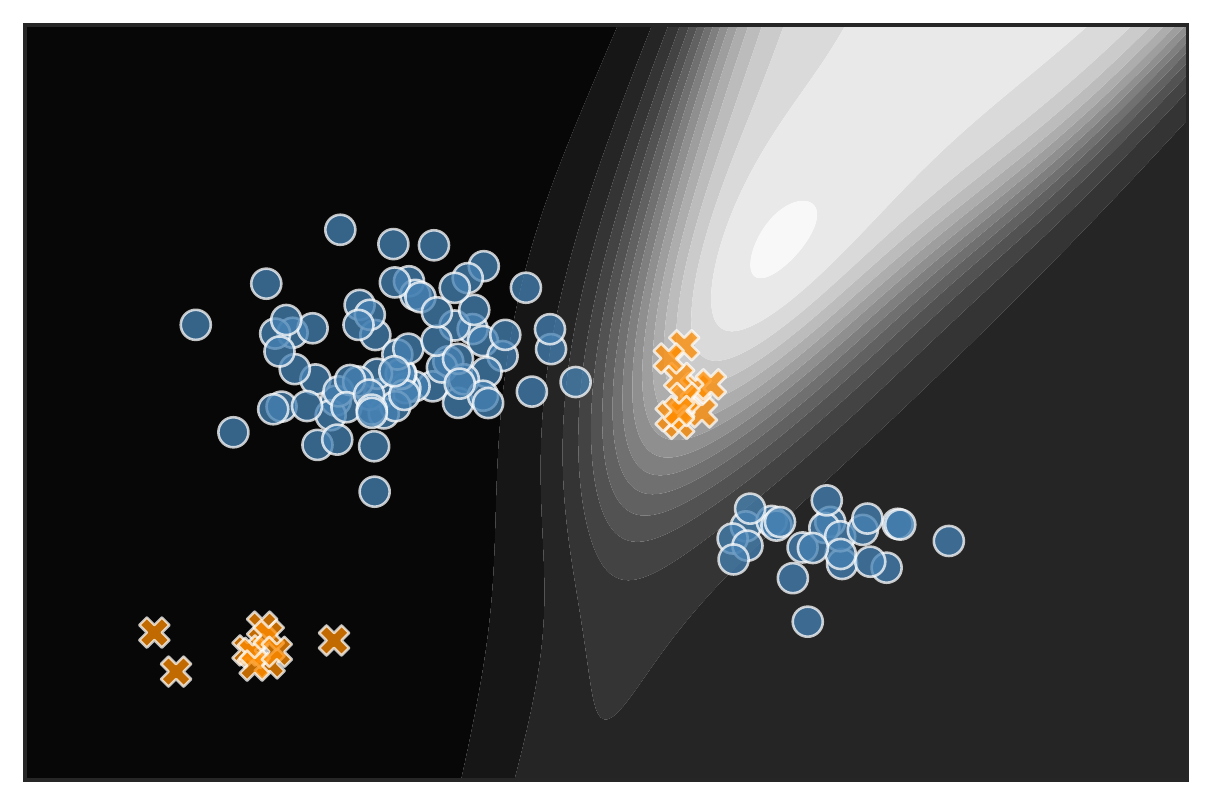}}
\subfigure[Semi-supervised LPUE]{\label{fig:toy_lpue}\includegraphics[width=0.329\linewidth]{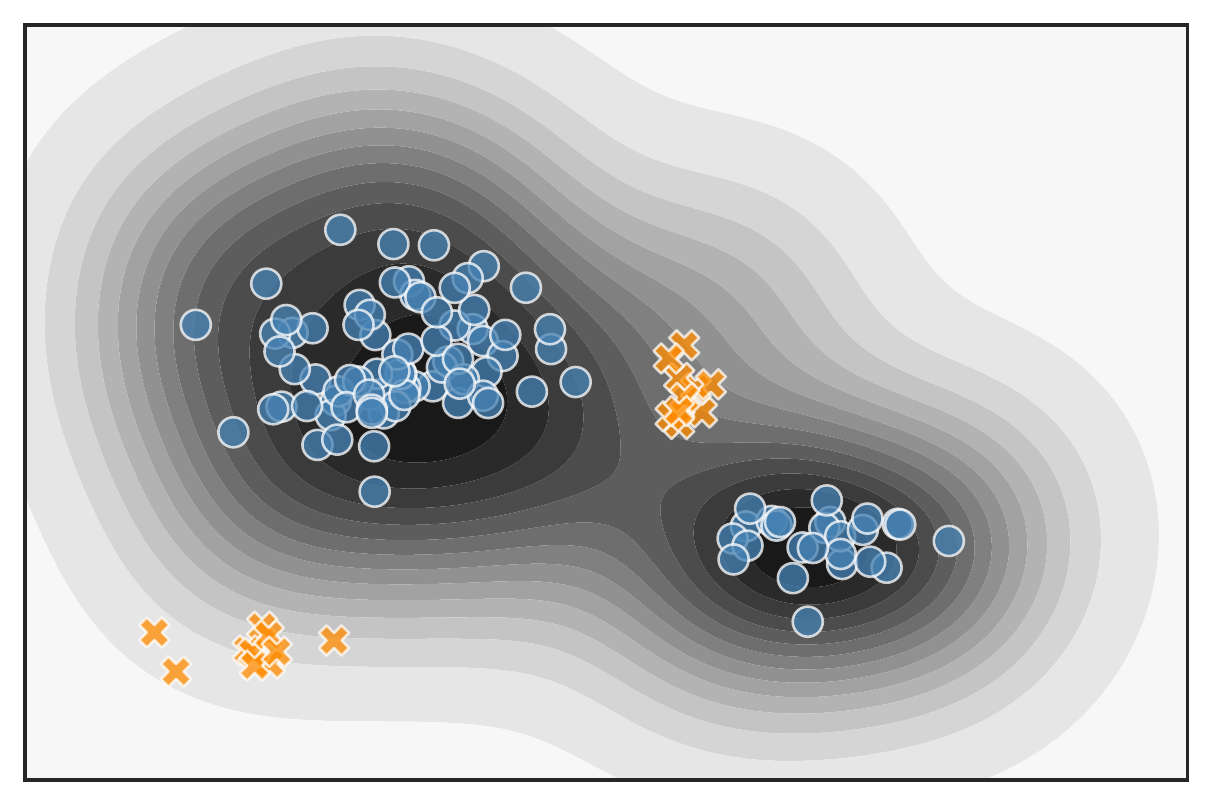}}
\subfigure[Semi-supervised AD (ours)]{\label{fig:toy_semisupervised_ad}\includegraphics[width=0.329\linewidth]{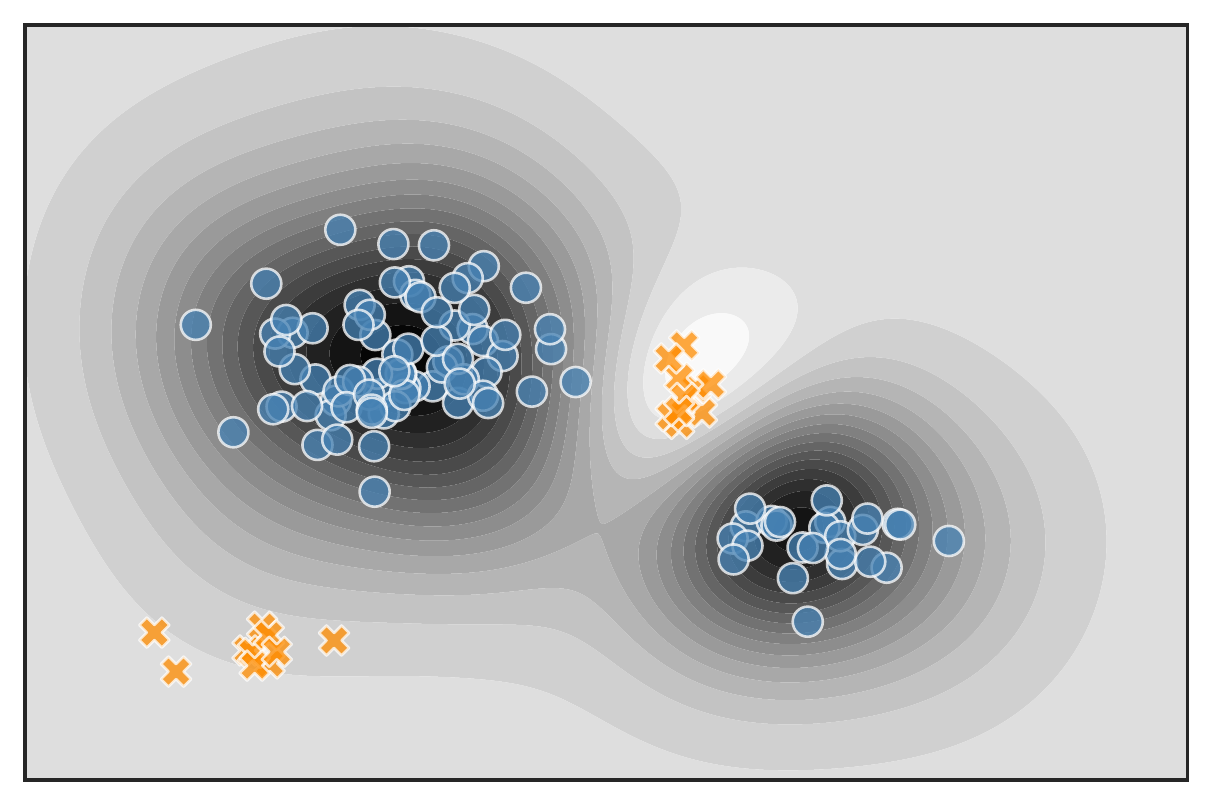}}
\end{center}
\vspace{-0.5em}
\caption{The need for semi-supervised anomaly detection: The training data (shown in (a)) consists of (mostly normal) unlabeled data (gray) as well as a few labeled normal samples (blue) and labeled anomalies (orange).
Figures (b)--(f) show the decision boundaries of the various learning paradigms at testing time along with novel anomalies that occur (bottom left in each plot).
Our semi-supervised AD approach takes advantage of all training data: unlabeled samples, labeled normal samples, as well as labeled anomalies. This strikes a balance between one-class learning and classification.
% The true normal data generating distribution is a mixture of two Gaussians.\\
% At testing time (shown in (b)--(f)) novel anomalies occur.
% A purely unsupervised model (b) does not take advantage of labeled training samples.
% A purely supervised model (c) overfits to the anomalies seen at training but fails to generalize to the novel anomalies.
% Learning from positive and unlabeled examples (LPUE) (d) improves over a fully unsupervised model using the labeled normal training samples, but still ignores the known anomalies, which are deemed normal.
% A semi-supervised classifier (e) makes a binary cluster assumption and also fails to generalize to the novel anomalies.
%Our semi-supervised AD approach (f) strikes a balance between one-class learning and classification.
}
\vspace{-0.5em}
\label{fig:toy_example}
\end{figure}
%%%%%%%%%%%%%%%%%%%%%%%%%%%%%%%%%%%%%%%%%%%%%%%%%%%%%%%%%%%%%%%%%%%%%%%%%%%%%%%%

% Introduction to AD and motivation for deep AD approaches
Anomaly detection (AD) \citep{chandola2009,pimentel2014} is the task of identifying unusual samples in data.
Typically AD methods attempt to learn a ``compact'' description of the data in an unsupervised manner assuming that most of the samples are normal (i.e., not anomalous).
For example, in one-class classification \citep{moya1993,scholkopf2001} the objective is to find a set of small measure which contains most of the data and samples not contained in that set are deemed anomalous.
Shallow unsupervised AD methods such as the One-Class SVM \citep{scholkopf2001,tax2004}, Kernel Density Estimation \citep{parzen1962,kim2012,vandermeulen2013}, or Isolation Forest \citep{liu2008} often require manual feature engineering to be effective on high-dimensional data and are limited in their scalability to large datasets.
These limitations have sparked great interest in developing novel \emph{deep} approaches to unsupervised AD \citep{erfani2016,zhai2016,chen2017,ruff2018,deecke2018,ruff2019,golan2018,pang2019,hendrycks2019,hendrycks2019using}.

% Motivation for *general* semi-supervised anomaly detection and problem formulation
Unlike the standard unsupervised AD setting, in many real-world applications one may also have access to some verified (i.e., labeled) normal or anomalous samples in addition to the unlabeled data.
Such samples could be hand labeled by a domain expert for instance.
This leads to a semi-supervised AD problem: given $n$ (mostly normal but possibly containing some anomalous contamination) unlabeled samples $\bm{x}_1, \ldots, \bm{x}_n$ and $m$ labeled samples $(\tilde{\bm{x}}_{1}, \tilde{y}_{1}), \ldots, (\tilde{\bm{x}}_{m}, \tilde{y}_{m})$, where $\tilde{y}={+}1$ and $\tilde{y}={-}1$ denote normal and anomalous samples respectively, the task is to learn a model that compactly characterizes the ``normal class.''

% Related work and clarification of "semi-supervised" term
The term \emph{semi-supervised anomaly detection} has been used to describe two different AD settings.
Most existing ``semi-supervised'' AD methods, both shallow \citep{munoz2010,blanchard2010,chandola2009} and deep \citep{song2017,akcay2018,chalapathy2019}, only incorporate the use of labeled normal samples but not labeled anomalies, i.e.~they are more precisely instances of \underline{L}earning from \underline{P}ositive (i.e., normal) and \underline{U}nlabeled \underline{E}xamples (LPUE) \citep{denis1998,zhang2008}.
A few works \citep{wang2005,liu2006,gornitz2013} have investigated the general semi-supervised AD setting where one also utilizes labeled anomalies, however existing deep approaches are domain or data-type specific \citep{ergen2017,kiran2018,min2018}.

Research on deep semi-supervised learning has almost exclusively focused on classification as the downstream task \citep{kingma2014b,rasmus2015,odena2016,dai2017,oliver2018}.
Such semi-supervised classifiers typically assume that similar points are likely to be of the same class, this is known as the \emph{cluster assumption} \citep{zhu2005,chapelle2009}.
This assumption, however, only holds for the ``normal class'' in AD, but is crucially invalid for the ``anomaly class'' since anomalies are not necessarily similar to one another.
% are by definition just not normal and do not have to be similar.
Instead, semi-supervised AD approaches must find a compact description of the normal class while also correctly discriminating the labeled anomalies \citep{gornitz2013}.
% Because of this, semi-supervised AD methods do not overfit to the labeled anomalies and generalize well to novel anomalies \citep{gornitz2013}.
Figure~\ref{fig:toy_example} illustrates the differences between various learning paradigms applied to AD on a toy example.

% Contribution
% 1. Information-theoretic perspective on Deep AD
% 2. Deep SAD method
% 3. Extensive experiments
We introduce \emph{Deep SAD} (Deep \underline{S}emi-supervised \underline{A}nomaly \underline{D}etection) in this work, an end-to-end deep method for general semi-supervised AD.
Our main contributions are the following:
\begin{itemize}
    \item We introduce Deep SAD, a generalization of the unsupervised Deep SVDD method \citep{ruff2018} to the semi-supervised AD setting.
    \item We present an information-theoretic framework for deep AD, which can serve as an interpretation of our Deep SAD method and similar approaches.
    \item We conduct extensive experiments in which we establish experimental scenarios for the general semi-supervised AD problem where we also introduce novel baselines.
\end{itemize}
% We show that our approach can be understood in information-theoretic terms as learning a latent distribution of low entropy for the normal data, with the anomalous distribution having a heavier tailed, higher entropy distribution.
% To do this we formulate an information-theoretic perspective on deep learning for AD.
% To the best of our knowledge it is the first method enabling the use of labeled samples in the field of deep AD for for general, high-dimensional data.

\section{An Information-theoretic View on Deep Anomaly Detection}
\label{sec:information_theory}

% Information Bottleneck (IB) principle
The study of the theoretical foundations of deep learning is an active and ongoing research effort \citep{montavon2011,tishby2015,cohen2016,eldan2016,neyshabur2017,raghu2017,zhang2016,achille2018,arora2018,belkin2018,wiatowski2018,lapuschkin2019}.
One important line of research that has emerged is rooted in information theory \citep{shannon1948}.
In the supervised classification setting where one has input variable $X$, latent variable $Z$ (e.g., the final layer of a deep network), and output variable $Y$ (i.e., the label), the well-known \emph{Information Bottleneck principle} \citep{tishby2000,tishby2015,shwartz2017,alemi2017,saxe2018} provides an explanation for representation learning as the trade-off between finding a minimal compression $Z$ of the input $X$ while retaining the informativeness of $Z$ for predicting the label $Y$.
Put formally, supervised deep learning seeks to minimize the mutual information $\mathcal{I}(X;Z)$ between the input $X$ and the latent representation $Z$ while maximizing the mutual information $\mathcal{I}(Z;Y)$ between $Z$ and the classification task $Y$, i.e.
\begin{equation}\label{eq:information_bottleneck}
    \min_{p(z | x)} \quad \mathcal{I}(X;Z) - \alpha \, \mathcal{I}(Z;Y),
\end{equation}
where $p(z|x)$ is modeled by a deep network and the hyperparameter $\alpha > 0$ controls the trade-off between compression (i.e., complexity) and classification accuracy.
% compression/complexity/regularization vs.~classification accuracy.
% minimal sufficient statistics

For unsupervised deep learning, due to the absence of labels $Y$ and thus the lack of a clear task, other information-theoretic learning principles have been formulated.
Of these, the \emph{Infomax principle} \citep{linsker1988,bell1995,hjelm2018} is one of the most prevalent and widely used principles.
In contrast to (\ref{eq:information_bottleneck}), the objective of Infomax is to \emph{maximize} the mutual information $\mathcal{I}(X;Z)$ between the data $X$ and its latent representation $Z$:
\begin{equation}\label{eq:infomax}
    \max_{p(z | x)} \quad \mathcal{I}(X;Z) + \beta \, \mathcal{R}(Z).
\end{equation}
This is typically done under some additional constraint or regularization $\mathcal{R}(Z)$ on the representation $Z$ with hyperparameter $\beta > 0$ to obtain statistical properties desired for some specific downstream task.
Examples where the Infomax principle has been applied include tasks such as independent component analysis \citep{bell1995}, clustering \citep{slonim2005,ji2018}, generative modeling \citep{chen2016,hoffman2016,zhao2017,alemi2018}, and unsupervised representation learning in general \citep{hjelm2018}.
%\citet{slonim2005} formulate the Blahut-Arimoto algorithm for clustering via mutual information maximization.

We observe that the Infomax principle has also been applied in previous deep representations for AD.
Most notably autoencoders \citep{rumelhart1986,hinton2006b}, which are the predominant approach to deep AD \citep{hawkins2002,sakurada2014,andrews2016,erfani2016,zhai2016,chen2017,chalapathy2019}, can be understood as implicitly maximizing the mutual information $\mathcal{I}(X;Z)$ via the reconstruction objective \citep{vincent2008} under some regularization of the latent code $Z$.
Choices for regularization include sparsity \citep{makhzani2013}, the distance to some latent prior distribution, e.g.~measured via the KL divergence \citep{kingma2013,rezende2014}, an adversarial loss \citep{makhzani2015}, or simply a bottleneck in dimensionality.
Such restrictions for AD share the idea that the latent representation of the normal data should be in some sense ``compact.''
%\footnote{Hybrid approaches separate these two aspects of ``information maximization'' and ``compact characterization'' by first learning representations (e.g., via autoencoder) and afterwards applying classical AD methods (e.g., the OC-SVM) to these representations in a second, separate step.}

As illustrated in Figure~\ref{fig:toy_example}, a supervised (or semi-supervised) classification approach to AD only learns to recognize anomalies similar to those seen during training, due to the class cluster assumption \citep{chapelle2009}.
However, \emph{anything} not normal is by definition an anomaly and thus anomalies do not have to be similar.
This makes supervised (or semi-supervised) classification learning principles such as (\ref{eq:information_bottleneck}) ill-defined for AD.
We instead build upon principle (\ref{eq:infomax}) to motivate a deep method for general semi-supervised AD, where we include the label information $Y$ through a novel representation learning regularization objective $\mathcal{R}(Z) = \mathcal{R}(Z; Y)$ that is based on entropy.
% constrain representations according to some desired statistical properties
% Our goal is to learn useful representations for the anomaly detection task.
% Without digging into the theory too much at this point, we think this principle may provide an underlying theoretical principle for deep AD which may guide the development of future methods.

\section{Deep Semi-supervised Anomaly Detection}
\label{sec:method}
In the following, we introduce \emph{Deep SAD}, a deep method for general semi-supervised AD.
To formulate our objective, we first briefly explain the unsupervised Deep SVDD method \citep{ruff2018} which we then generalize to the semi-supervised AD setting.
%To formulate our objective, we first show that the unsupervised Deep SVDD method \citep{ruff2018} can be interpreted in terms of an entropy minimization objective on the latent representation.
%We then generalize the method to the semi-supervised AD setting.

\subsection{Unsupervised Deep SVDD and Entropy Minimization}
\label{sec:deepSVDD}

For input space $\mathcal{X} \subseteq \mathbb{R}^D$ and output space $\mathcal{Z} \subseteq \mathbb{R}^d$, let $\phi(\cdot \, ; \mathcal{W}) : \mathcal{X} \to \mathcal{Z}$ be a neural network with $L$ hidden layers and corresponding set of weights $\mathcal{W} = \{\bm{W}^1, \ldots, \bm{W}^L \}$.
The objective of Deep SVDD is to train the neural network $\phi$ to learn a transformation that minimizes the volume of a data-enclosing hypersphere in output space $\mathcal{Z}$ centered on a predetermined point $\bm{c}$.
Given $n$ (unlabeled) training samples $\bm{x}_1, \ldots, \bm{x}_n \in \mathcal{X}$, the \emph{One-Class Deep SVDD} objective is
\begin{equation}\label{eq:deepSVDD_oc}
  \min_{\mathcal{W}} \quad \frac{1}{n} \sum_{i=1}^n \Vert \phi(\bm{x}_i; \mathcal{W}) - \bm{c} \Vert^2 + \frac{\lambda}{2} \sum_{\ell=1}^L \Vert \bm{W}^\ell \Vert_F^2, \quad \lambda > 0.
\end{equation}
Penalizing the mean squared distance of the mapped samples to the hypersphere center $\bm{c}$ forces the network to extract those common factors of variation which are most stable within the dataset.
As a consequence normal data points tend to get mapped near the hypersphere center, whereas anomalies are mapped further away \citep{ruff2018}.
%Minimizing the volume of the sphere enforces this learning process.
The second term is a standard weight decay regularizer.

Deep SVDD is optimized via SGD using backpropagation.
For initialization, \citet{ruff2018} first pre-train an autoencoder and then initialize the weights $\mathcal{W}$ of the network $\phi$ with the converged weights of the encoder.
After initialization, the hypersphere center $\bm{c}$ is set as the mean of the network outputs obtained from an initial forward pass of the data.
Once the network is trained, the anomaly score for a test point $\bm{x}$ is given by the distance from $\phi(\bm{x}; \mathcal{W})$ to the center of the hypersphere:
\begin{equation}\label{eq:deepSVDD_score}
  s(\bm{x}) = \Vert \phi(\bm{x}; \mathcal{W}) - \bm{c} \Vert.
\end{equation}

We now argue that Deep SVDD may not only be interpreted in geometric terms as minimum volume estimation \citep{scott2006}, but also in probabilistic terms as entropy minimization over the latent distribution.
For a latent random variable $Z$ with covariance $\Sigma$, pdf $p(\bm{z})$, and support $\mathcal{Z} \subseteq \mathbb{R}^d$, we have the following bound on entropy
\begin{equation}
    \mathcal{H}(Z) = \mathbb{E}[-\log p(Z)] = - \int_{\mathcal{Z}} p(\bm{z}) \log p(\bm{z}) \, \mathrm{d}\bm{z} \leq \frac{1}{2} \log((2 \pi e)^d \det \Sigma),
\end{equation}
which holds with equality iff $Z$ is jointly Gaussian \citep{cover2012}.
Assuming the latent distribution $Z$ follows an isotropic Gaussian, $Z \sim N(\bm{\mu}, \sigma^2 I)$ with $\sigma > 0$, we get
\begin{equation}
    \mathcal{H}(Z) = \frac{1}{2} \log((2 \pi e)^d \det \sigma^2 I) = \frac{1}{2} \log((2 \pi e \sigma^2)^d \cdot 1)  = \frac{d}{2} (1 + \log(2 \pi \sigma^2)) \propto \log \sigma^2,
\end{equation}
i.e.~for a fixed dimensionality $d$, the entropy of $Z$ is proportional to its log-variance.

Now observe that the Deep SVDD objective (\ref{eq:deepSVDD_oc}) (disregarding weight decay regularization) is equivalent to minimizing the empirical variance and thus minimizes an upper bound on the entropy of a latent Gaussian.
Since the Deep SVDD network is pre-trained on an autoencoding objective that implicitly maximizes the mutual information $\mathcal{I}(X;Z)$ \citep{vincent2008}, we may interpret Deep SVDD as following the Infomax principle (\ref{eq:infomax}) with the additional ``compactness'' objective that the latent distribution should have minimal entropy.

\subsection{Deep SAD}
\label{sec:deepSAD}

We now introduce our method for deep semi-supervised anomaly detection: \emph{Deep SAD}.
Assume that, in addition to the $n$ unlabeled samples $\bm{x}_1, \ldots, \bm{x}_n \in \mathcal{X}$ with $\mathcal{X} \subseteq \mathbb{R}^D$, we also have access to $m$ labeled samples $(\tilde{\bm{x}}_{1}, \tilde{y}_{1}), \ldots, (\tilde{\bm{x}}_{m}, \tilde{y}_{m}) \in \mathcal{X} \times \mathcal{Y}$ with $\mathcal{Y} = \{{-}1, {+}1 \}$ where $\tilde{y}={+}1$ denotes known normal samples and $\tilde{y}={-}1$ known anomalies.
We define our \emph{Deep SAD} objective as follows:
\begin{equation}\label{eq:deepSAD}
	\min_{\mathcal{W}} \quad \frac{1}{n{+}m} \sum_{i=1}^n \Vert \phi(\bm{x}_i; \mathcal{W}) - \bm{c} \Vert^2 + \frac{\eta}{n{+}m} \sum_{j=1}^m \left( \Vert \phi(\tilde{\bm{x}}_{j}; \mathcal{W}) - \bm{c} \Vert^2 \right)^{\tilde{y}_{j}} + \frac{\lambda}{2} \sum_{\ell=1}^L \Vert \bm{W}^\ell \Vert_F^2.
\end{equation}
We employ the same loss term as Deep SVDD for the unlabeled data in our Deep SAD objective and thus recover Deep SVDD (\ref{eq:deepSVDD_oc}) as the special case when there is no labeled training data available ($m=0$).
In doing this we also incorporate the assumption that most of the unlabeled data is normal.

For the labeled data, we introduce a new loss term that is weighted via the hyperparameter $\eta > 0$ which controls the balance between the labeled and the unlabeled term.
Setting $\eta > 1$ puts more emphasis on the labeled data whereas $\eta < 1$ emphasizes the unlabeled data.
For the labeled normal samples ($\tilde{y}={+}1$), we also impose a quadratic loss on the distances of the mapped points to the center $\bm{c}$, thus intending to overall learn a latent distribution which concentrates the normal data.
Again, one might consider $\eta > 1$ to emphasize labeled normal over unlabeled samples.
For the labeled anomalies ($\tilde{y}={-}1$) in contrast, we penalize the \emph{inverse} of the distances such that anomalies must be mapped further away from the center.\footnote{To ensure numerical stability, we add a machine epsilon (\texttt{eps} $\sim 10^{-6}$) to the denominator of the inverse.}
Note that this is in line with the common assumption that anomalies are not concentrated \citep{scholkopf2001learning,steinwart2005}.
In our experiments we found that simply setting $\eta = 1$ yields a consistent and substantial performance improvement. A sensitivity analysis on $\eta$ is in Section \ref{sec:sensitivtiy_analysis}.

We define the Deep SAD anomaly score again by the distance of the mapped point to the center $c$ as given in Eq.~(\ref{eq:deepSVDD_score}) and optimize our Deep SAD objective (\ref{eq:deepSAD}) via SGD using backpropagation.
We provide a summary of the Deep SAD optimization procedure and further details in Appendix \ref{sec:app_optimization}.

In addition to the inverse squared norm loss we experimented with several other losses including the negative squared norm loss, negative robust losses, and the hinge loss.
The negative squared norm loss, which is unbounded from below, resulted in an ill-posed optimization problem and caused optimization to diverge.
Negative robust losses, such as the Hampel loss, introduce one or more scale parameters which are difficult to select or optimize in conjunction with the changing representation learned by the network. 
Like \citet{ruff2018}, we observed that the hinge loss was difficult to optimize and resulted in poorer performance.
The inverse squared norm loss instead is bounded from below and smooth, which are crucial properties for losses used in deep learning \citep{goodfellow2016}, and ultimately performed the best while remaining conceptually simple.
% Inverse squared norm loss beneficial properties: bounded, smooth, yet conceptually simple with few hyperparameters.

Following our insights on the connection between Deep SVDD and entropy minimization from Section \ref{sec:deepSVDD}, we may interpret our Deep SAD objective as modeling the latent distribution of normal data, $Z^+ = Z|\{Y{=}{+}{1}\}$, to have \emph{low entropy}, and the latent distribution of anomalies, $Z^- = Z|\{Y{=}{-}{1}\}$, to have \emph{high entropy}.
Minimizing the distances to the center $\bm{c}$ (i.e., minimizing the empirical variance) for the mapped points of labeled normal samples ($\tilde{y}={+}1$) induces a latent distribution with low entropy for the normal data.
In contrast, penalizing low variance via the inverse squared norm loss for the mapped points of labeled anomalies ($\tilde{y}={-}1$) induces a latent distribution with high entropy for the anomalous data.
That is, the network must attempt to map known anomalies to some heavy-tailed distribution.
We argue that such a model better captures the nature of anomalies, which can be thought of as being generated from an infinite mixture of distributions that are different from the normal data distribution, indubitably a distribution that has high entropy.
Our objective notably does not impose any cluster assumption on the anomaly-generating distribution $X|\{Y{=}{-}{1}\}$ as is typically made in supervised or semi-supervised classification approaches \citep{zhu2005,chapelle2009}.
We can express this interpretation in terms of principle (\ref{eq:infomax}) with an entropy regularization objective on the latent distribution:
\begin{equation}\label{eq:infomax_semi}
    \max_{p(z | x)} \quad \mathcal{I}(X;Z) + \beta \, (\mathcal{H}(Z^-) - \mathcal{H}(Z^+)).
\end{equation}
To maximize the mutual information $\mathcal{I}(X;Z)$, Deep SAD also relies on autoencoder pre-training \citep{vincent2008,ruff2018}.

\section{Experiments}
\label{sec:experiments}
We evaluate Deep SAD on MNIST, Fashion-MNIST, and CIFAR-10 as well as on classic AD benchmark datasets.
We compare to shallow, hybrid, as well as deep unsupervised, semi-supervised and supervised competitors.
We refer to other recent works \citep{ruff2018,golan2018,hendrycks2019} for further comparisons between unsupervised deep AD methods.\footnote{Our code is available at: \url{https://github.com/lukasruff/Deep-SAD-PyTorch}}

\subsection{Competing Methods}
We consider the OC-SVM \citep{scholkopf2001} and SVDD \citep{tax2004} with Gaussian kernel (which in this case are equivalent), Isolation Forest \citep{liu2008}, and KDE \citep{parzen1962} for shallow unsupervised baselines.
For deep unsupervised competitors, we consider well-established (convolutional) autoencoders and the state-of-the-art unsupervised Deep SVDD method \citep{ruff2018}.
To avoid confusion, we note again that some literature \citep{song2017,chalapathy2019} refer to the methods above as being ``semi-supervised'' if they are trained on only labeled normal samples.
For general semi-supervised AD approaches that also take advantage of labeled anomalies, we consider the state-of-the-art shallow SSAD method \citep{gornitz2013} with Gaussian kernel.
%, which is a semi-supervised extension of kernel SVDD \citep{tax2004}.
As mentioned earlier, there are no deep competitors for general semi-supervised AD that are applicable to general data types.
To get a comprehensive comparison we therefore introduce a novel \emph{hybrid SSAD} baseline that applies SSAD to the latent codes of autoencoder models.
Such hybrid methods have demonstrated solid performance improvements over their raw feature counterparts on high-dimensional data \citep{erfani2016,nicolau2016}.
We also include such hybrid variants for all unsupervised shallow competitors.
To also compare to a deep semi-supervised learning method that targets classification as the downstream task, we add the well-known Semi-Supervised Deep Generative Model (SS-DGM) \citep{kingma2014b} where we use the latent class probability estimate (normal vs. anomalous) as the anomaly score.
To complete the full learning spectrum, we also include a fully supervised deep classifier trained on the binary cross-entropy loss.

In our experiments we deliberately grant the shallow and hybrid methods an unfair advantage by selecting their hyperparameters to maximize AUC on a subset (10\%) of the test set to minimize hyperparameter selection issues.
To control for architectural effects between the deep methods, we always use the same (LeNet-type) deep networks.
Full details on network architectures and hyperparameter selection can be found in Appendices \ref{sec:app_architectures} and \ref{sec:app_competitors}.
Due to space constraints, in the main text we only report results for methods which showed competitive performance and defer results for the underperforming methods in Appendix \ref{sec:app_full_results}.

\subsection{Experimental Scenarios on MNIST, Fashion-MNIST, and CIFAR-10}
\label{sec:exp_scenarios}

\textbf{Semi-supervised anomaly detection setup \;} MNIST, Fashion-MNIST, and CIFAR-10 all have ten classes from which we derive ten AD setups on each dataset following previous works \citep{ruff2018,chalapathy2018,golan2018}.
In every setup, we set one of the ten classes to be the normal class and let the remaining nine classes represent anomalies.
We use the original training data of the respective normal class as the unlabeled part of our training set.
Thus we start with a clean AD setting that fulfills the assumption that most (in this case all) unlabeled samples are normal.
% This leads to unlabeled training data sizes of $n \approx 6\,000$ for MNIST and Fashion-MNIST, and $n = 5\,000$ for CIFAR-10 per AD setup.
The training data of the respective nine anomaly classes then forms the data pool from which we draw anomalies for training to create different scenarios.
We compute the commonly used AUC measure on the original respective test sets using ground truth labels to make a quantitative comparison, i.e.~$\tilde{y}={+}1$ for the normal class and $\tilde{y}={-}1$ for the respective nine anomaly classes.
We rescale pixels to $[0,1]$ via min-max feature scaling as the only data pre-processing step.
% Training data (semi-supervised): unlabeled data (mostly normal, yet also polluted) + labeled data (anomalies or normal examples)
% Test data: normal examples + anomalies from known anomaly classes + novel anomalies

\textbf{Experimental scenarios \;} We examine three scenarios in which we vary the following three experimental parameters: (i) the ratio of labeled training data $\lab$, (ii) the ratio of pollution $\pol$ in the unlabeled training data with (unknown) anomalies, and (iii) the number of anomaly classes $\klab$ included in the labeled training data.

\textbf{(i) Adding labeled anomalies \;} In this scenario, we investigate the effect that including labeled anomalies during training has on detection performance to see the benefit of a general semi-supervised AD approach over other paradigms.
To do this we increase the ratio of labeled training data $\lab = m{/}(n{+}m)$ by adding more and more known anomalies $\tilde{\bm{x}}_{1}, \ldots, \tilde{\bm{x}}_{m}$ with $\tilde{y}_{j} = -1$ to the training set.
The labeled anomalies are sampled from one of the nine anomaly classes ($\klab = 1$).
For testing, we then consider all nine remaining classes as anomalies, i.e.~there are eight novel classes at testing time.
We do this to simulate the unpredictable nature of anomalies.
For the unlabeled part of the training set, we keep the training data of the respective normal class, which we leave unpolluted in this experimental setup, i.e.~$\pol = 0$.
% Note that the unlabeled part of the training set is unpolluted for now, i.e.~$\pol = 0$.
We iterate this training set generation process per AD setup always over all the nine respective anomaly classes and report the average results over the ten AD setups $\times$ nine anomaly classes, i.e.~over 90 experiments per labeled ratio $\lab$.

\textbf{(ii) Polluted training data \;} Here we investigate the robustness of the different methods to an increasing pollution ratio $\pol$ of the training set with unlabeled anomalies.
To do so we pollute the unlabeled part of the training set with anomalies drawn from all nine respective anomaly classes in each AD setup.
We fix the ratio of labeled training samples at $\lab = 0.05$ where we again draw samples only from $\klab = 1$ anomaly class in this scenario.
We repeat this training set generation process per AD setup over all the nine respective anomaly classes and report the average results over the resulting 90 experiments per pollution ratio $\pol$.
We hypothesize that learning from labeled anomalies in a semi-supervised AD approach alleviates the negative impact pollution has on detection performance since similar unknown anomalies in the unlabeled data might be detected.

\textbf{(iii) Number of known anomaly classes \;} In the last scenario, we compare the detection performance at various numbers of known anomaly classes.
In scenarios (i) and (ii), we always sample labeled anomalies only from one out of the nine anomaly classes ($\klab = 1$).
In this scenario, we now increase the number of anomaly classes $\klab$ included in the labeled part of the training set.
Since we have a limited number of anomaly classes (nine) in each AD setup, we expect the supervised classifier to catch up at some point.
We fix the overall ratio of labeled training examples again at $\lab = 0.05$ and consider a pollution ratio of $\pol = 0.1$ for the unlabeled training data in this scenario.
We repeat this training set generation process for ten seeds in each of the ten AD setups and report the average results over the resulting 100 experiments per number $\klab$.
For each seed, the $\klab$ classes are drawn uniformly at random from the nine respective anomaly classes.

%%%%%%%%%%%%%%%%%%%%%%%%%%%%%%%%%%%%%%%%%%%%%%%%%%%%%%%%%%%%%%%%%%%%%%%%%%%%%%%%
\begin{figure}[th]
\begin{center}
%\vspace{-0.5em}
\includegraphics[width=\linewidth]{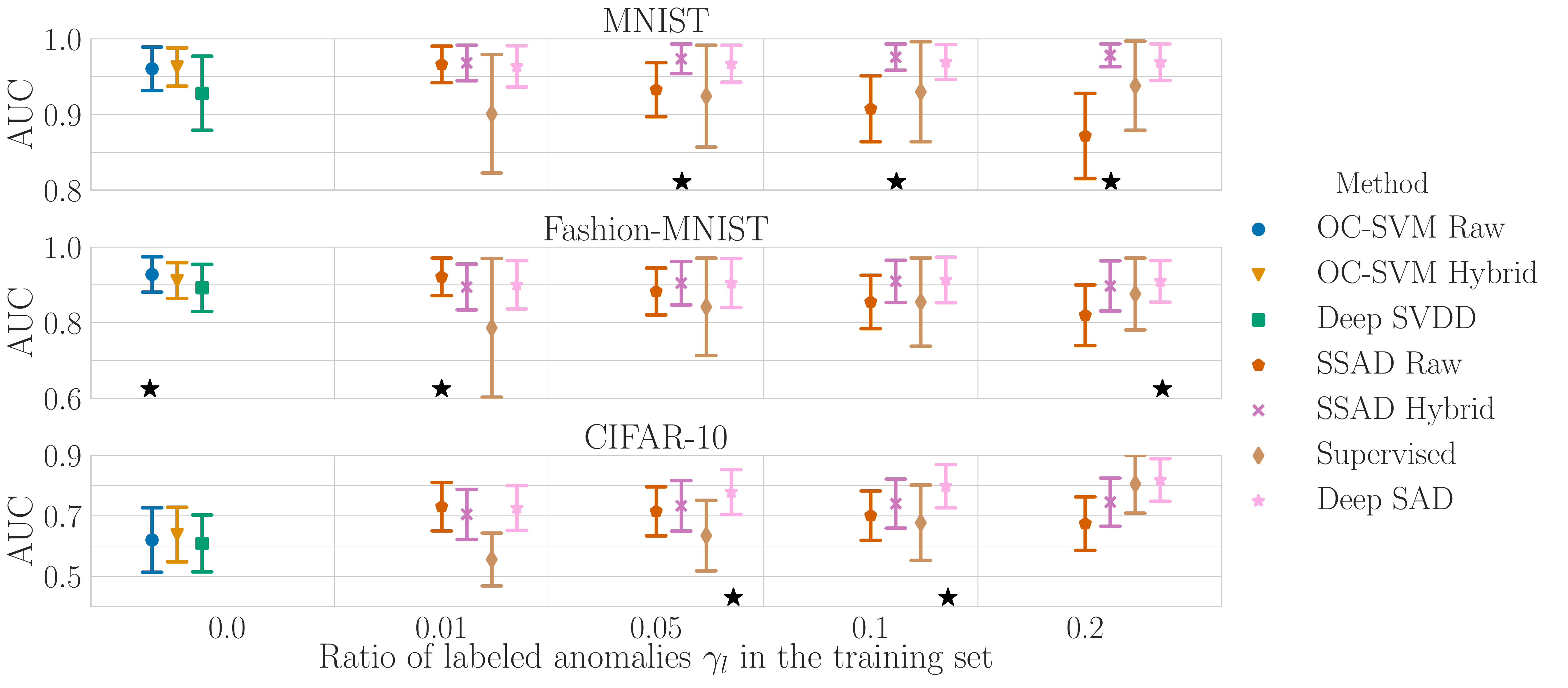}
\end{center}
\vspace{-1.0em}
\caption{Results of scenario (i), where we increase the ratio of labeled anomalies $\lab$ in the training set. We report avg.~AUC with st.~dev.~over 90 experiments at various ratios $\lab$. A ``$\star$'' indicates a statistically significant ($\alpha=0.05$) difference between the 1\textsuperscript{st} and 2\textsuperscript{nd} best method.}
\label{fig:1_known}
%\vspace{-0.5em}
\end{figure}
%%%%%%%%%%%%%%%%%%%%%%%%%%%%%%%%%%%%%%%%%%%%%%%%%%%%%%%%%%%%%%%%%%%%%%%%%%%%%%%%

%%%%%%%%%%%%%%%%%%%%%%%%%%%%%%%%%%%%%%%%%%%%%%%%%%%%%%%%%%%%%%%%%%%%%%%%%%%%%%%%
\begin{figure}[th]
\begin{center}
\vspace{-1.5em}
\includegraphics[width=\linewidth]{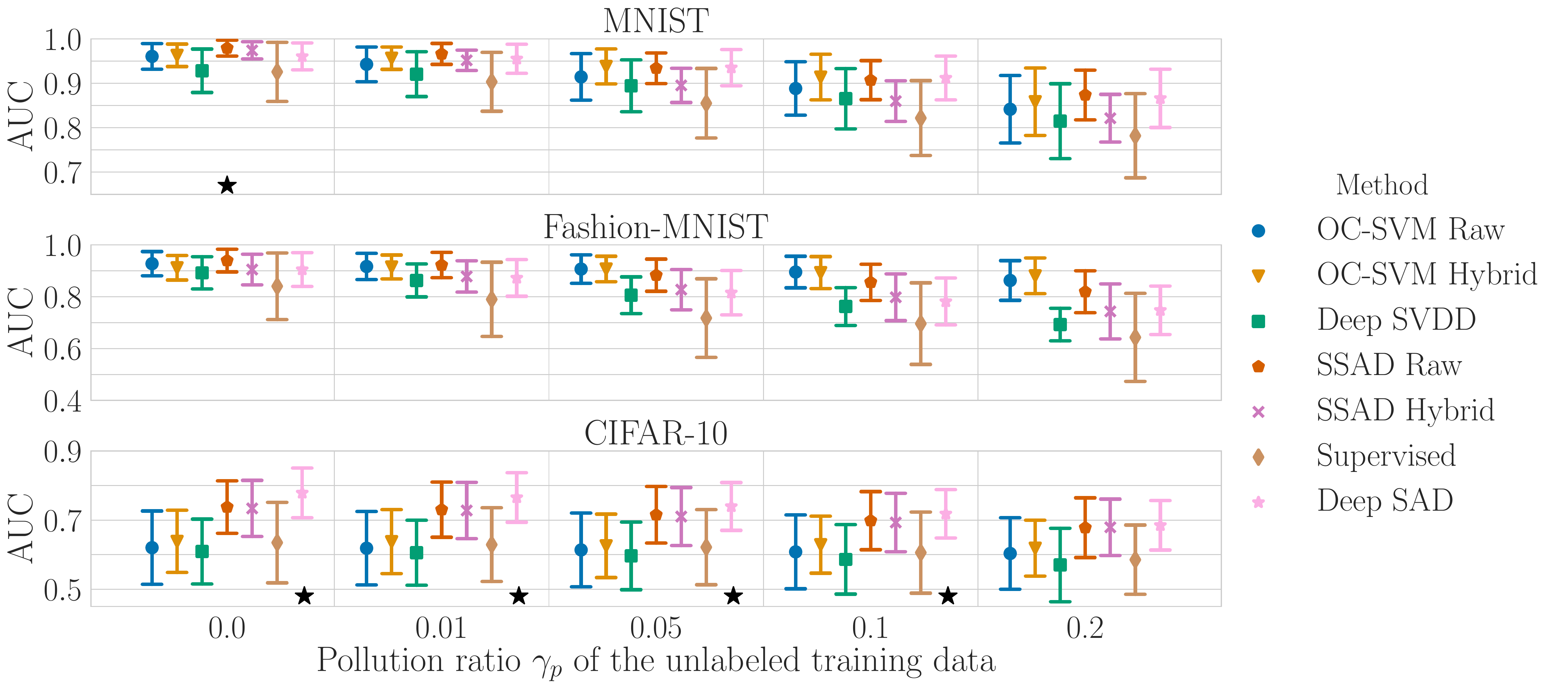}
\end{center}
\vspace{-1.0em}
\caption{Results of scenario (ii), where we pollute the unlabeled part of the training set with (unknown) anomalies. We report avg.~AUC with st.~dev.~over 90 experiments at various ratios $\pol$. A ``$\star$'' indicates a statistically significant ($\alpha=0.05$) difference between the 1\textsuperscript{st} and 2\textsuperscript{nd} best method.}
\label{fig:pollution}
\vspace{-0.5em}
\end{figure}
%%%%%%%%%%%%%%%%%%%%%%%%%%%%%%%%%%%%%%%%%%%%%%%%%%%%%%%%%%%%%%%%%%%%%%%%%%%%%%%%

%%%%%%%%%%%%%%%%%%%%%%%%%%%%%%%%%%%%%%%%%%%%%%%%%%%%%%%%%%%%%%%%%%%%%%%%%%%%%%%%
\begin{figure}[t]
\begin{center}
\vspace{-0.5em}
\includegraphics[width=\linewidth]{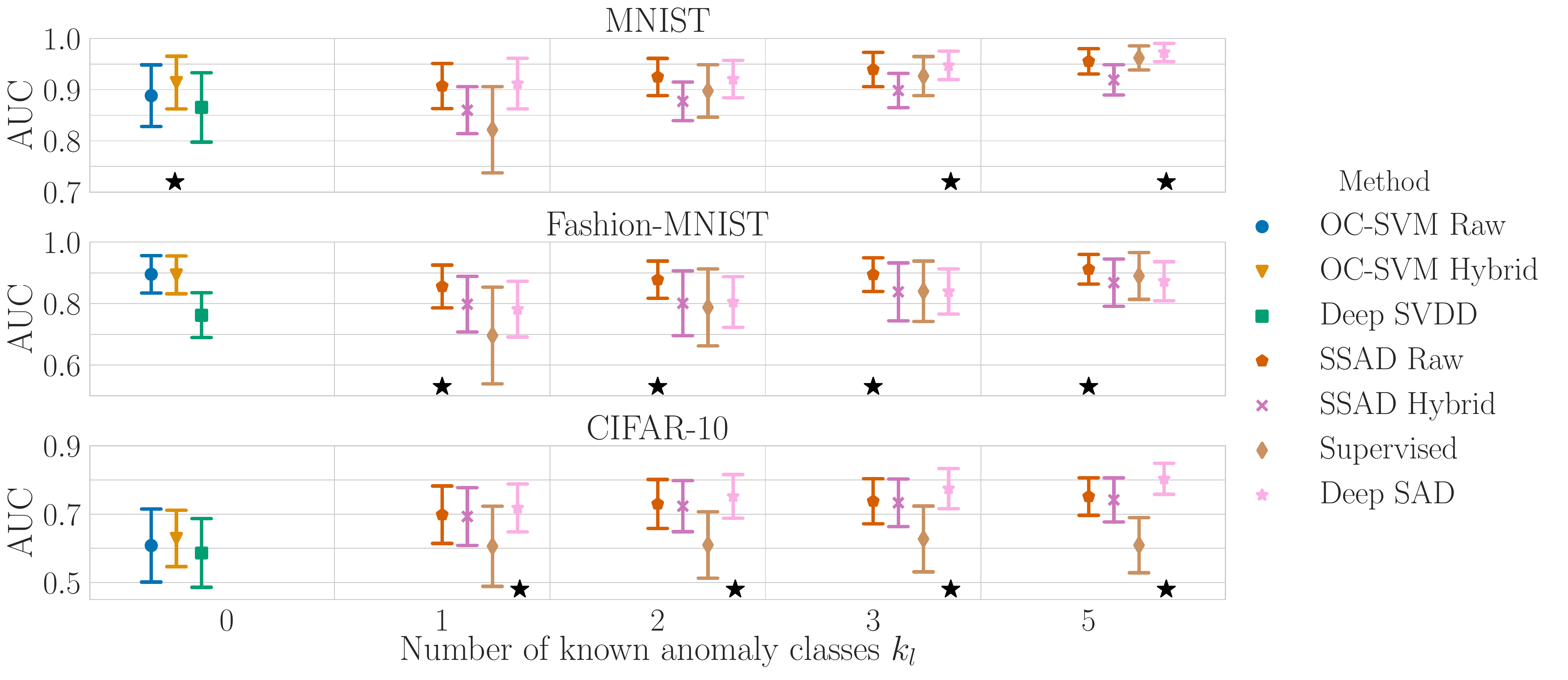}
\end{center}
\vspace{-1.0em}
\caption{Results of scenario (iii), where we increase the number of anomaly classes $\klab$ included in the labeled training data. We report avg.~AUC with st.~dev.~over 100 experiments for various $\klab$. A ``$\star$'' indicates a statistically significant ($\alpha=0.05$) difference between the 1\textsuperscript{st} and 2\textsuperscript{nd} best method.}
\label{fig:n_known}
\vspace{-1.5em}
\end{figure}
%%%%%%%%%%%%%%%%%%%%%%%%%%%%%%%%%%%%%%%%%%%%%%%%%%%%%%%%%%%%%%%%%%%%%%%%%%%%%%%%

\textbf{Results \;} The results of scenarios (i)--(iii) are shown in Figures \ref{fig:1_known}--\ref{fig:n_known}.
In addition to the avg.~AUC with st.~dev., we report the outcome of Wilcoxon signed-rank tests \citep{wilcoxon1945} applied to the first and second best performing method to indicate statistically significant ($\alpha=0.05$) differences in performance.
Figure~\ref{fig:1_known} demonstrates the benefit of our semi-supervised approach to AD especially on the most complex CIFAR-10 dataset, where Deep SAD performs best.
% On the less complex MNIST and Fashion-MNIST datasets, the unsupervised detectors already establish a strong baseline.
Figure~\ref{fig:1_known} moreover confirms that a supervised classification approach is vulnerable to novel anomalies at testing time when only little labeled training data is available.
In comparison, Deep SAD generalizes to novel anomalies while also taking advantage of the labeled examples.
Note that our novel hybrid SSAD baseline also performs well.
Figure~\ref{fig:pollution} shows that the detection performance of all methods decreases with increasing data pollution.
Deep SAD proves to be most robust again especially on CIFAR-10.
Finally, Figure~\ref{fig:n_known} shows that the more diverse the labeled anomalies in the training set, the better the detection performance becomes.
We can again see that the supervised method is very sensitive to the number of anomaly classes but catches up at some point as suspected. This does not occur with CIFAR-10, however, where $\gamma_l = 0.05$ labeled training samples seems to be insufficient for classification.
Overall, we see that Deep SAD is particularly beneficial on the more complex data.

\vspace{1.5em}
\subsection{Sensitivity analysis}
\label{sec:sensitivtiy_analysis}

%%%%%%%%%%%%%%%%%%%%%%%%%%%%%%%%%%%%%%%%%%%%%%%%%%%%%%%%%%%%%%%%%%%%%%%%%%%%%%%%
\begin{wrapfigure}{R}{0.5\textwidth}
    \centering
    \vspace{-1.6em}
    \includegraphics[width=\linewidth]{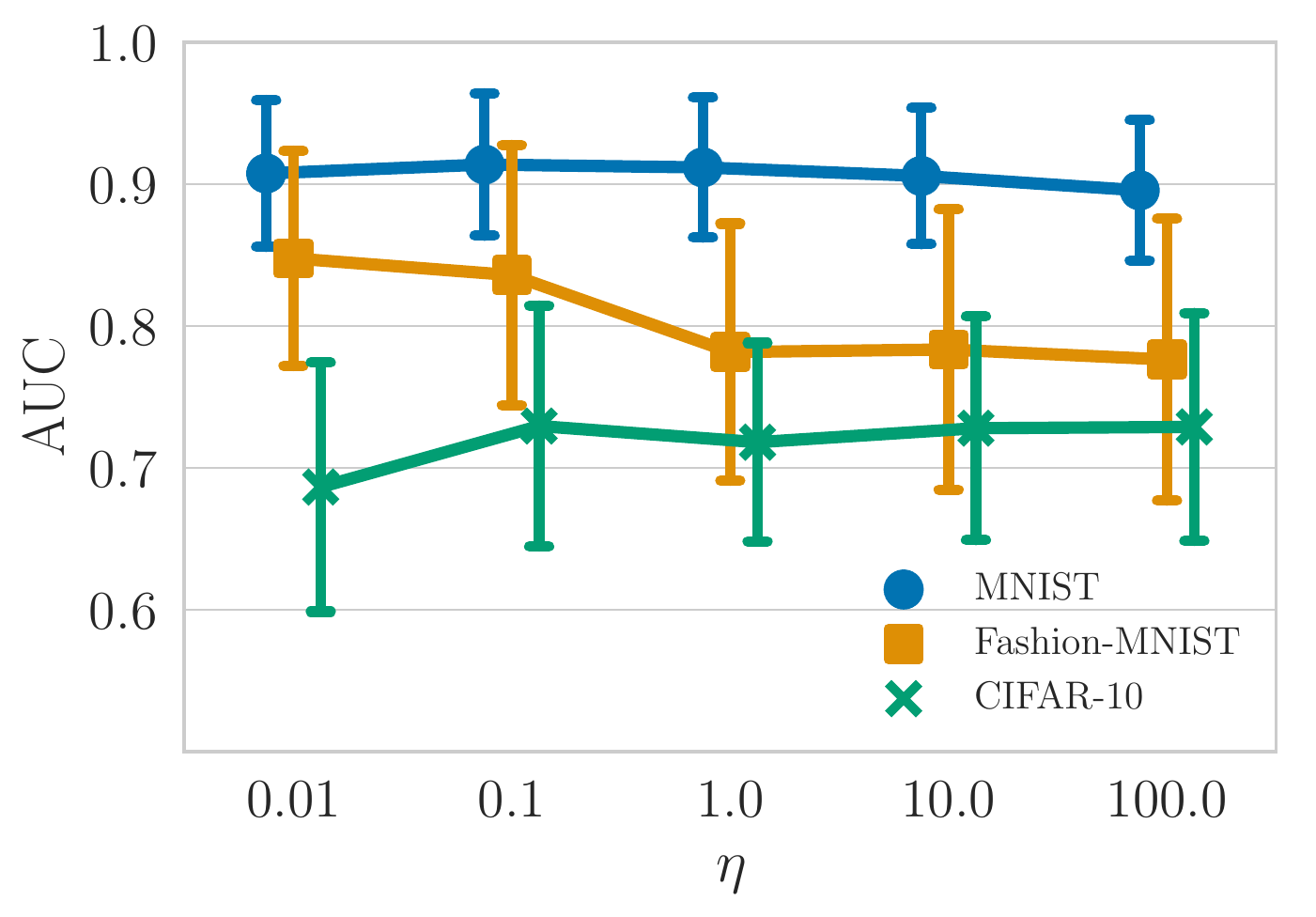}
    \vspace{-0.85em}
    \caption{Deep SAD sensitivity analysis w.r.t.~$\eta$. We report avg.~AUC with st.~dev.~over 90 experiments for various values of hyperparameter $\eta$.}
    \label{fig:eta_sensitivity}
    \vspace{-0.5em}
\end{wrapfigure}
%%%%%%%%%%%%%%%%%%%%%%%%%%%%%%%%%%%%%%%%%%%%%%%%%%%%%%%%%%%%%%%%%%%%%%%%%%%%%%%%
We run Deep SAD experiments on the ten AD setups described above on each dataset for $\eta \in \{10^{-2}, \ldots, 10^{2}\}$ to analyze the sensitivity of Deep SAD with respect to the hyperparameter $\eta>0$.
In this analysis, we set the experimental parameters to their default, $\gamma_l = 0.05$, $\gamma_p = 0.1$, and $k_l = 1$, and again iterate over all nine anomaly classes in every AD setup.
The results shown in Figure~\ref{fig:eta_sensitivity} suggest that Deep SAD is fairly robust against changes of the hyperparameter $\eta$.

In addition, we run experiments under the same experimental settings while varying the dimension $d \in \{2^4, \ldots, 2^9 \}$ of the output space $\mathcal{Z} \subseteq \mathbb{R}^d$ to infer the sensitivity of Deep SAD with respect to the representation dimensionality, where we keep $\eta = 1$.
The results are given in Figure \ref{fig:dim_sensitivity} in Appendix \ref{sec:app_exp_scenarios}. 
There we also compare to our hybrid SSAD baseline, which was the strongest competitor.
Interestingly we observe that detection performance increases with dimension $d$, converging to an upper bound in performance.
This suggests that one would want to set $d$ large enough to have sufficiently high mutual information $\mathcal{I}(X;Z)$ before compressing to a compact characterization.

\subsection{Classic Anomaly Detection Benchmark Datasets}
\label{sec:exp_classic}

In a final experiment, we also examine the detection performance of the various methods on some well-established AD benchmark datasets \citep{rayana2016}.
We run these experiments to evaluate the deep versus the shallow approaches on non-image datasets that are rarely considered in deep AD literature.
Here we observe that the shallow kernel methods seem to have a slight edge on the relatively small, low-dimensional benchmarks.
Nonetheless, Deep SAD proves competitive and the small differences observed might be explained by the advantage we grant the shallow methods in their hyperparameter selection.
We give the full details and results in Appendix \ref{sec:app_exp_classic}.

Our results and other recent works \citep{ruff2018,golan2018,hendrycks2019} overall demonstrate that deep methods are especially superior on complex data with hierarchical structure.
Unlike other deep approaches \citep{ergen2017,kiran2018,min2018,deecke2018,golan2018}, however, our Deep SAD method is not domain or data-type specific. Due to its good performance using both deep and shallow networks we expect Deep SAD to extend well to other data types.

\section{Conclusion and Future Work}
\label{sec:conclusion}
In this work we introduced Deep SAD, a deep method for general semi-supervised anomaly detection.
Our method is a generalization of the unsupervised Deep SVDD method \citep{ruff2018} to the semi-supervised setting.
The results of our experimental evaluation suggest that general semi-supervised anomaly detection should always be preferred whenever some labeled information on both normal samples or anomalies is available.

Moreover, we formulated an information-theoretic framework for deep anomaly detection based on the Infomax principle. 
Using this framework, we interpreted our method as minimizing the entropy of the latent distribution for normal data and maximizing the entropy of the latent distribution for anomalous data.
We introduced this framework with the aim of forming a basis for new methods as well as rigorous theoretical analyses in the future, e.g.~studying deep anomaly detection under the rate-distortion curve \citep{alemi2018}.

%%%%%%%%%%%%%%%%%%%%%%%%%%%%%%%%%%%%%%%%%%%%%%%%%%%%%%%%%%%%%%%%%%%%%%%%%%%%%%%%
% Notes
% 
% Future Work:
% * Theoretical analysis of the information-theoretic view on deep anomaly detection.
% * Informative negative sampling and Active Learning strategies
% * Explainability

% Informative negative sampling and explainability
% * Observation: the more similar the anomaly class is to the normal class, the greater the improvement in detection performance.
% * Hypothesis: the closer the labeled examples are to the boundary, the more informative they are.
% * Make connection to active learning
% * Add explainability heatmaps (gradient $\times$ input; LRP) after training with different known anomaly classes. (In MNIST experiment 3 vs.~all, the anomaly classes 6, 8, 9 are most informative.)
%%%%%%%%%%%%%%%%%%%%%%%%%%%%%%%%%%%%%%%%%%%%%%%%%%%%%%%%%%%%%%%%%%%%%%%%%%%%%%%%

% If you'd like to, you may include a section for author contributions as is done in many journals. 
% This is optional and at the discretion of the authors.

\subsubsection*{Acknowledgments}
% Use unnumbered third level headings for the acknowledgments. 
% All acknowledgments, including those to funding agencies, go at the end of the paper.
LR acknowledges support by the German Ministry of Education and Research (BMBF) in the project ALICE III (01IS18049B).
MK and RV acknowledge support by the German Research Foundation (DFG) award KL 2698/2-1 and by the German Ministry of Education and Research (BMBF) awards 031L0023A, 01IS18051A, and 031B0770E. 
AB is grateful for support by the National Research Foundation of Singapore, STEE-SUTD Cyber Security Laboratory, and the Ministry of Education, Singapore, under its program MOE2016-T2-2-154.
NG acknowledges support by the German Ministry of Education and Research (BMBF) through the Berlin Center for Machine Learning (01IS18037I).
KRM acknowledges partial financial support by the German Ministry of Education and Research (BMBF) under grants 01IS14013A-E, 01IS18025A, 01IS18037A, 01GQ1115 and 01GQ0850; Deutsche Forschungsgesellschaft (DFG) under grant Math+, EXC 2046/1, project-ID 390685689, and by the Technology Promotion (IITP) grant funded by the Korea government (No. 2017-0-00451, No. 2017-0-01779).

\bibliography{main.bib}
\bibliographystyle{iclr2020_conference}

\clearpage
\appendix

%%%%%%%%%%%%%%%%%%%%%%%%%%%%%%%%%%%%%%%%%%%%%%%%%%%%%%%%%%%%%%%%%%%%%%%%%%%%%%%%
% Supplementary Material
%%%%%%%%%%%%%%%%%%%%%%%%%%%%%%%%%%%%%%%%%%%%%%%%%%%%%%%%%%%%%%%%%%%%%%%%%%%%%%%%

\section{Additional Results on MNIST, Fashion-MNIST, and CIFAR-10}
\label{sec:app_exp_scenarios}

\subsection{Sensitivity Analysis w.r.t~Representation Dimensionality}
%%%%%%%%%%%%%%%%%%%%%%%%%%%%%%%%%%%%%%%%%%%%%%%%%%%%%%%%%%%%%%%%%%%%%%%%%%%%%%%%
\begin{figure}[ht]
\begin{center}
\vspace{-0.5em}
\includegraphics[width=\linewidth]{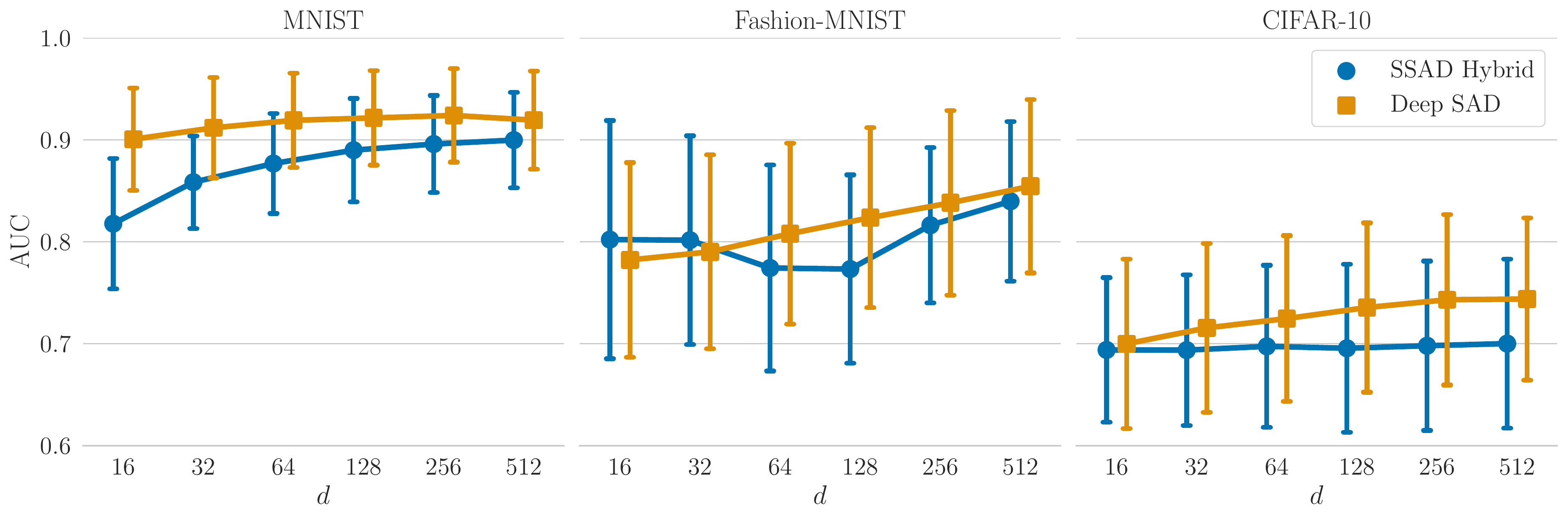}
\end{center}
\vspace{-0.5em}
\caption{Sensitivity analysis w.r.t.~the network representation dimensionality $d$ for our Deep SAD method and the closest competitor hybrid SSAD. We report avg.~AUC with st.~dev.~over 90 experiments for various values of $d$.}
\label{fig:dim_sensitivity}
\vspace{-0.5em}
\end{figure}
%%%%%%%%%%%%%%%%%%%%%%%%%%%%%%%%%%%%%%%%%%%%%%%%%%%%%%%%%%%%%%%%%%%%%%%%%%%%%%%%

\subsection{AUC Scatterplots of Best vs.~Second Best Methods on CIFAR-10}
\begin{figure}[ht]
\centering
\subfigure[$\lab = 0.01$]{\label{fig:1_known_scatter_a}\includegraphics[width=0.329\linewidth]{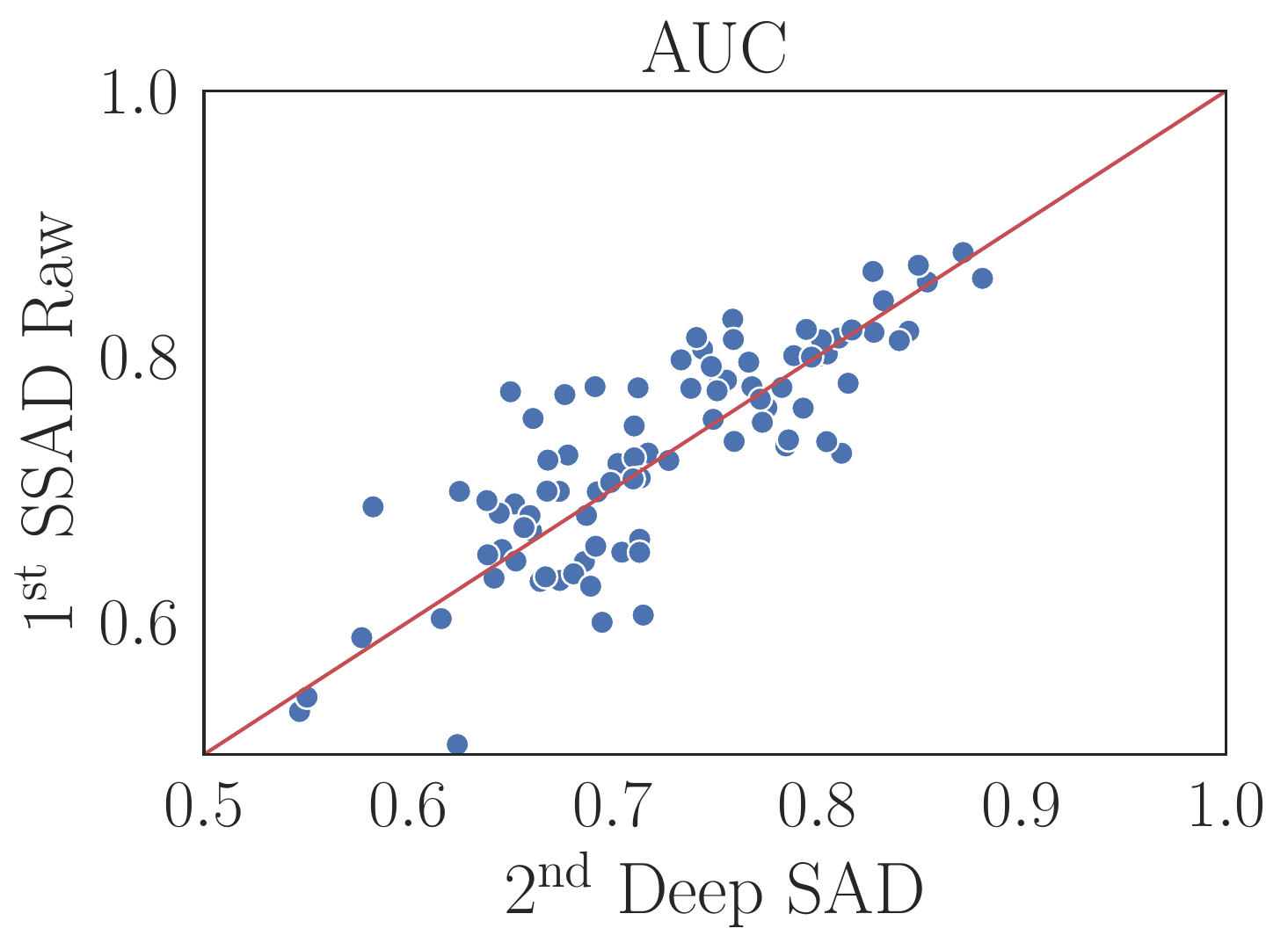}}
\subfigure[$\lab = 0.05$]{\label{fig:1_known_scatter_b}\includegraphics[width=0.329\linewidth]{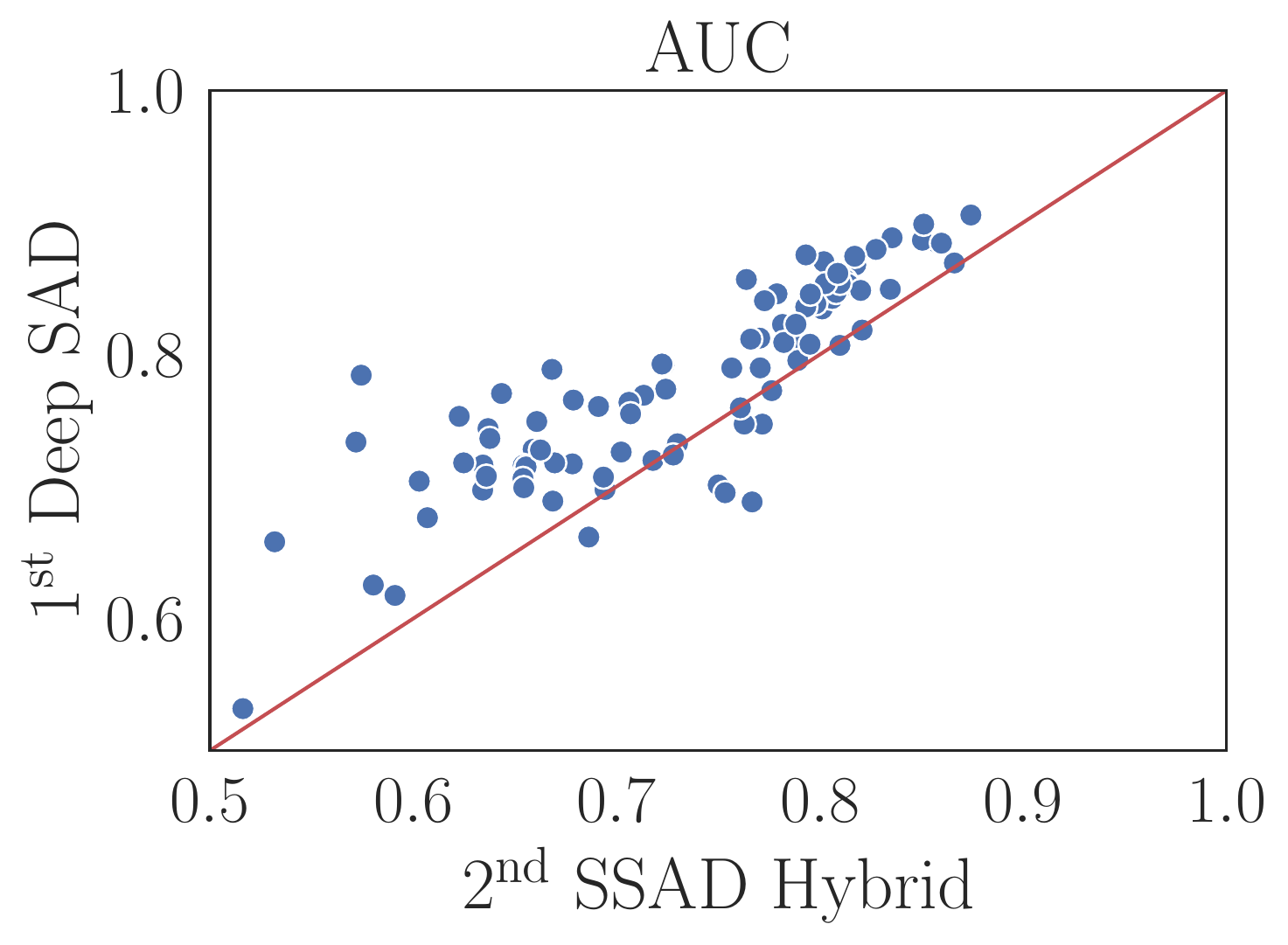}}
\subfigure[$\lab = 0.1$]{\label{fig:1_known_scatter_c}\includegraphics[width=0.329\linewidth]{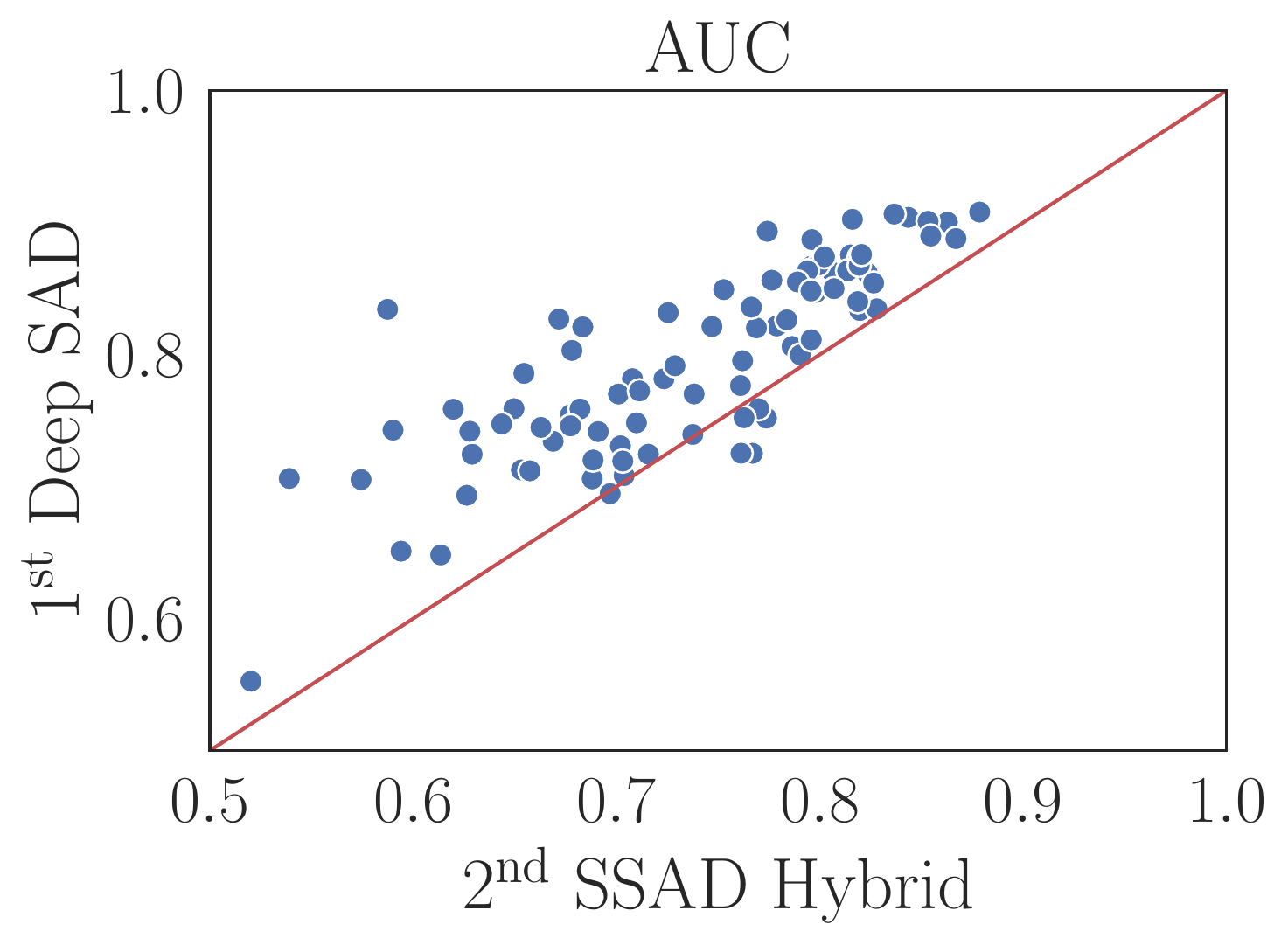}}
\subfigure[$\lab = 0.2$]{\label{fig:1_known_scatter_d}\includegraphics[width=0.329\linewidth]{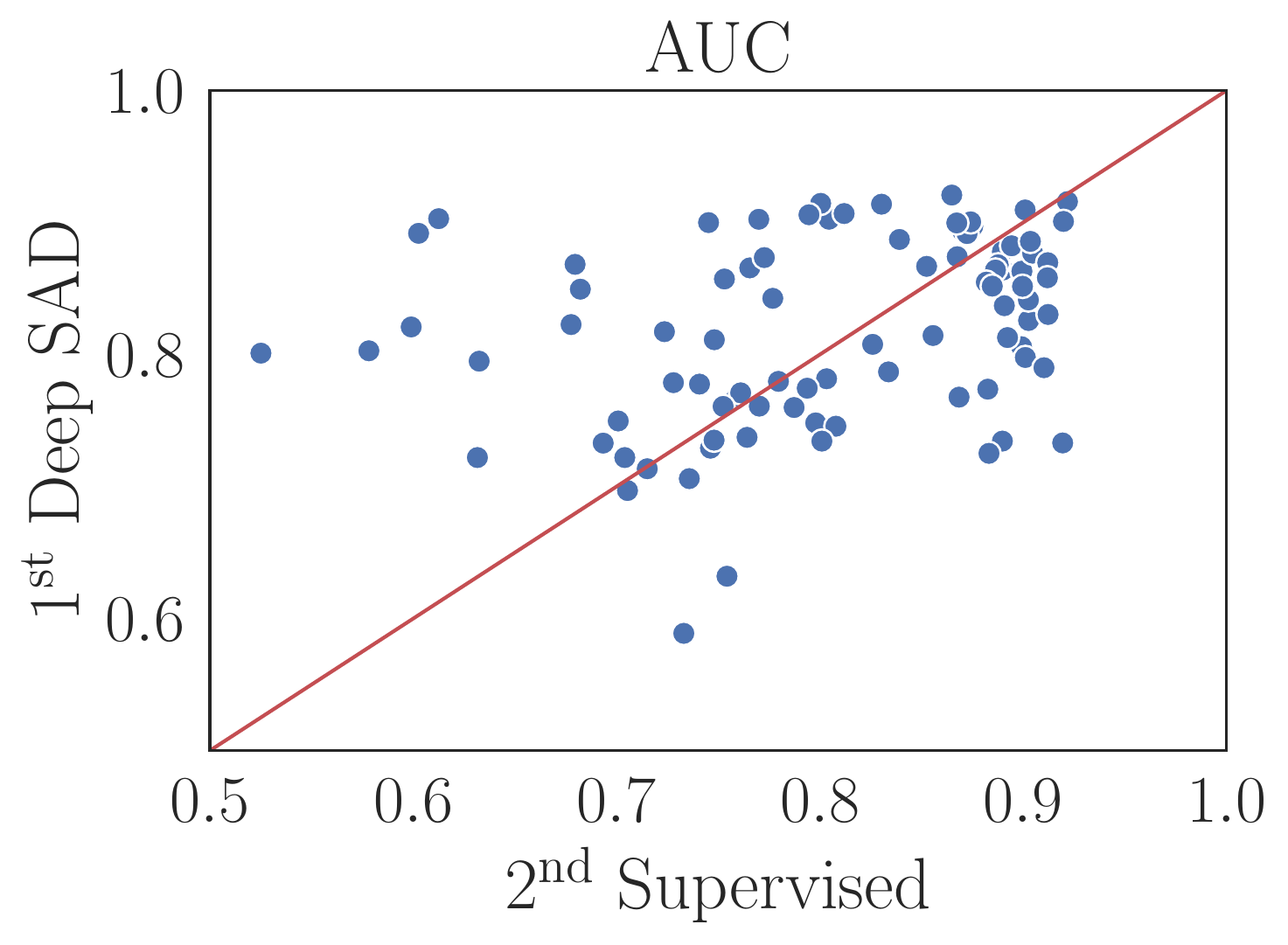}}
\caption{AUC scatterplots of best (1\textsuperscript{st}) vs.~second best (2\textsuperscript{nd}) performing methods in experimental scenario (i) on CIFAR-10, where we increase the ratio of labeled anomalies $\lab$ in the training set.}
\label{fig:1_known_scatter}
\end{figure}

\begin{figure}[ht]
\centering
\subfigure[$\pol = 0$]{\label{fig:pollution_scatter_a}\includegraphics[width=0.329\linewidth]{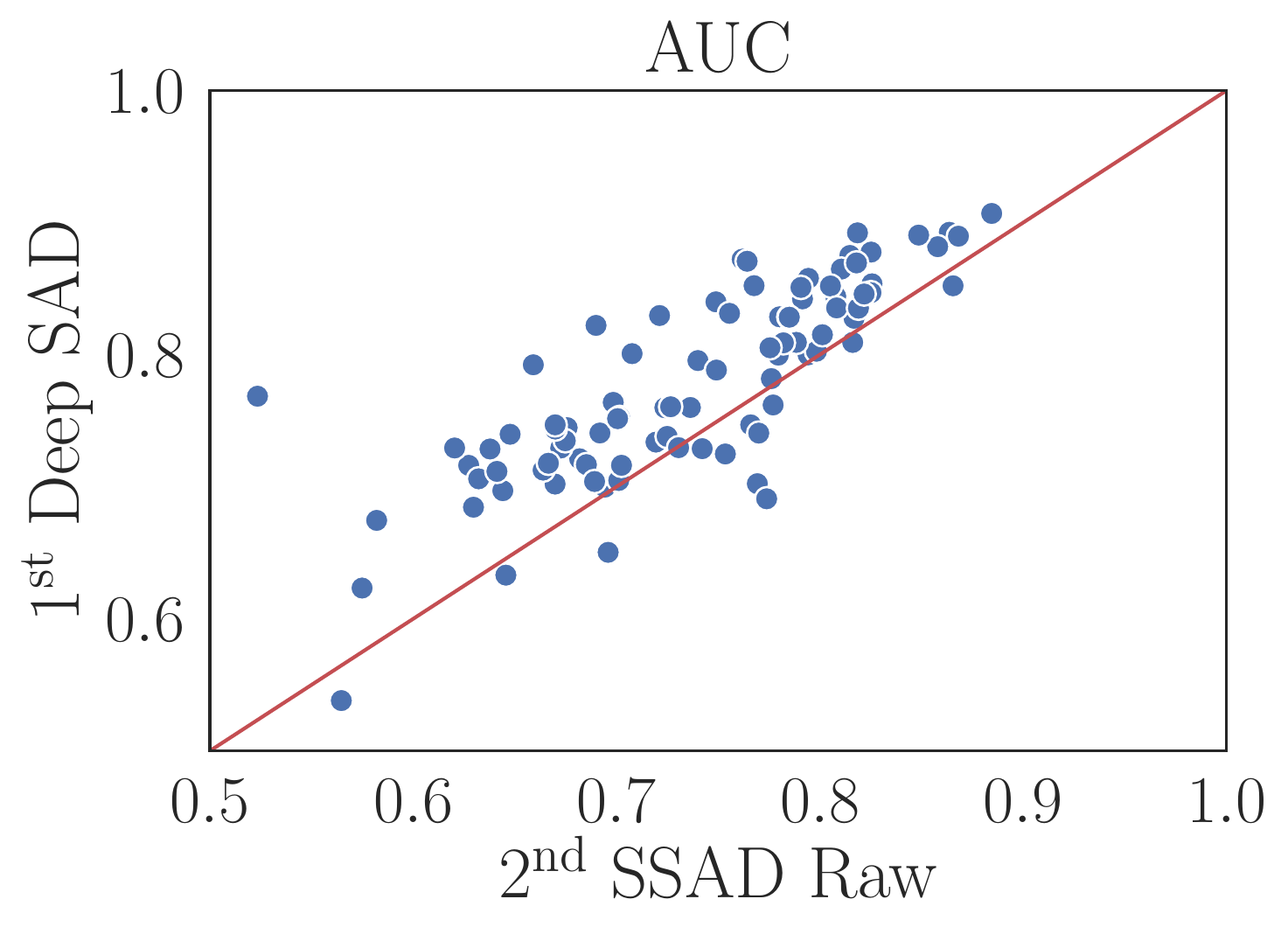}}
\subfigure[$\pol = 0.01$]{\label{fig:pollution_scatter_b}\includegraphics[width=0.329\linewidth]{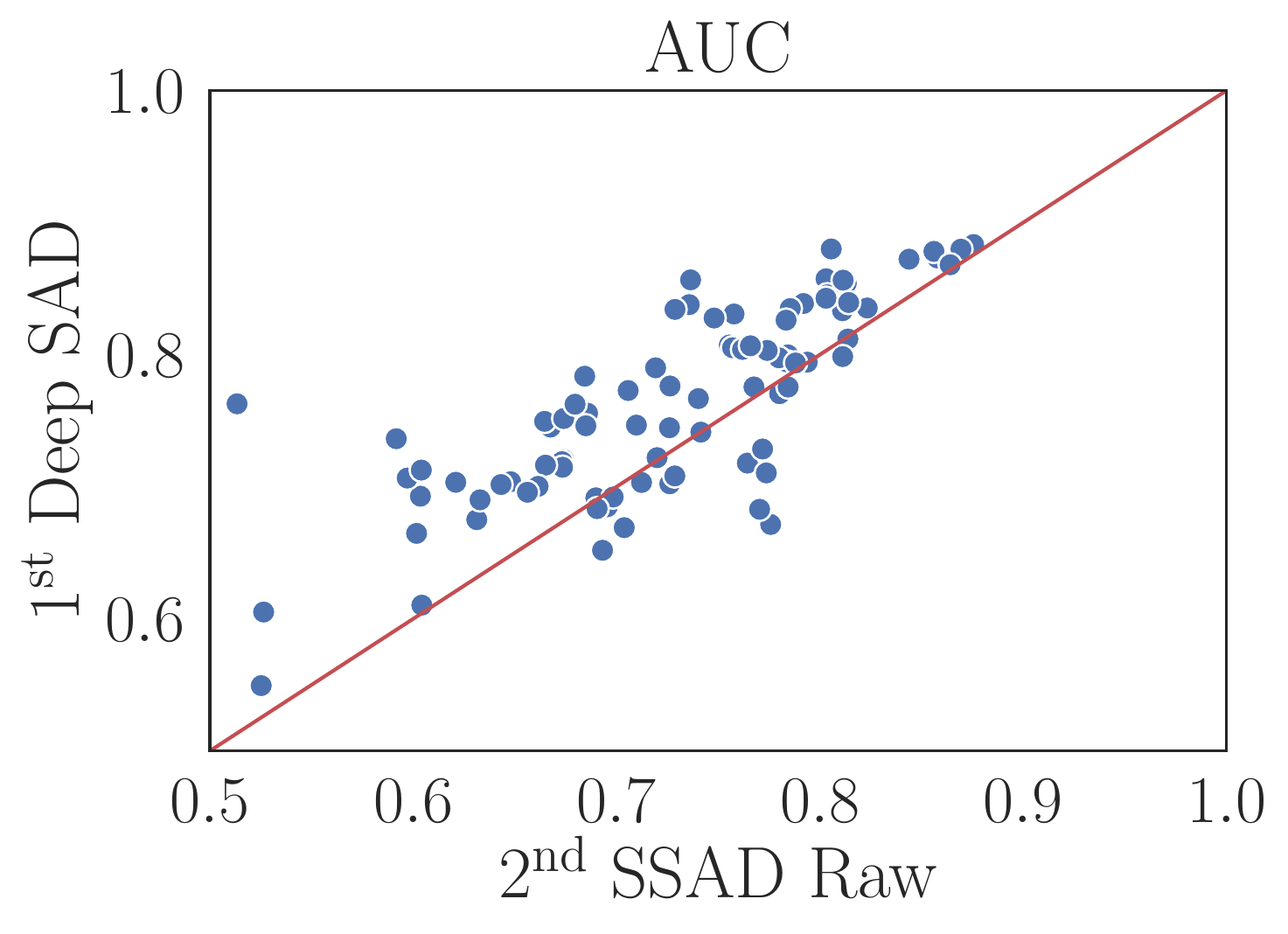}}
\subfigure[$\pol = 0.05$]{\label{fig:pollution_scatter_c}\includegraphics[width=0.329\linewidth]{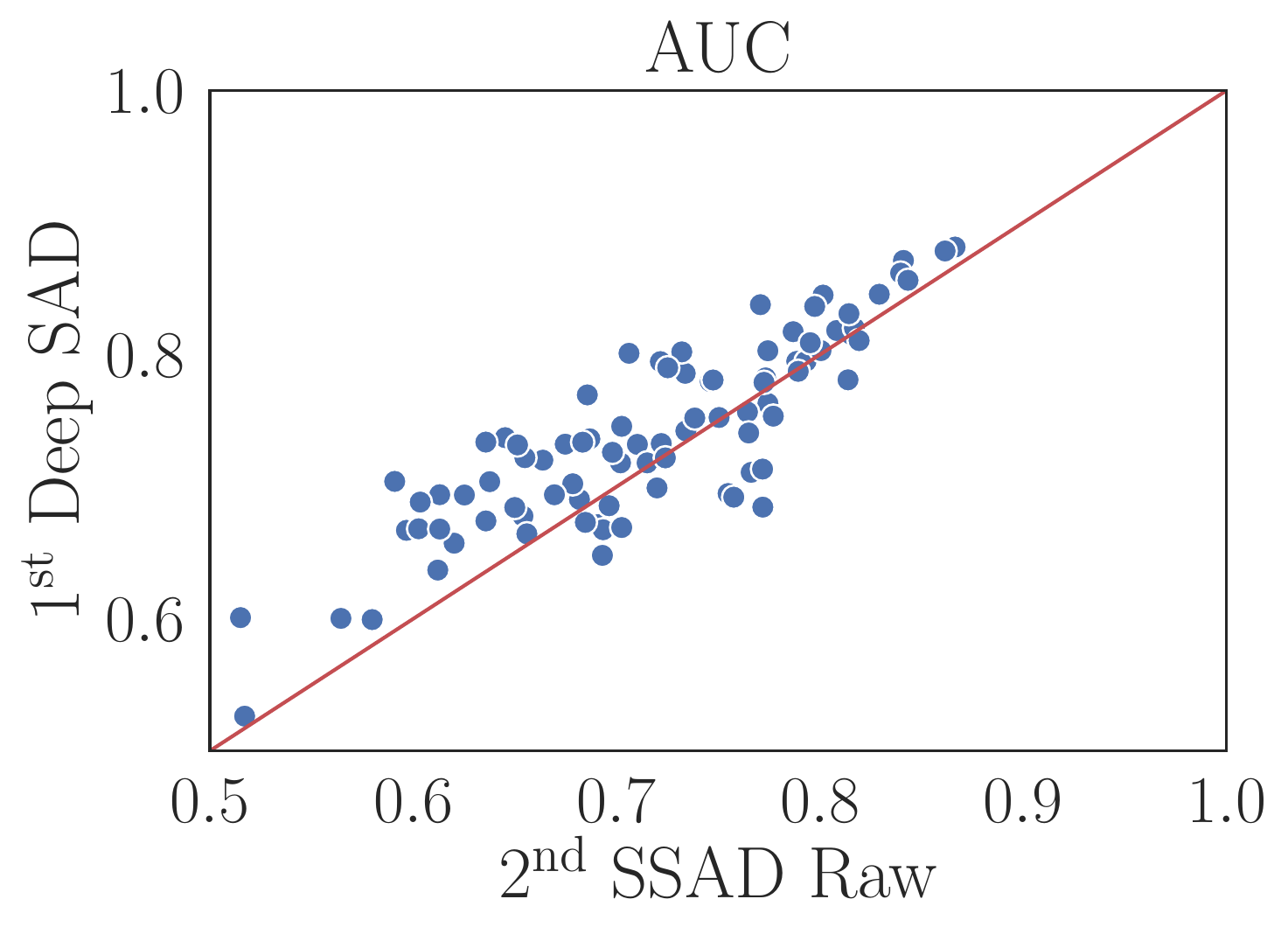}}
\subfigure[$\pol = 0.1$]{\label{fig:pollution_scatter_d}\includegraphics[width=0.329\linewidth]{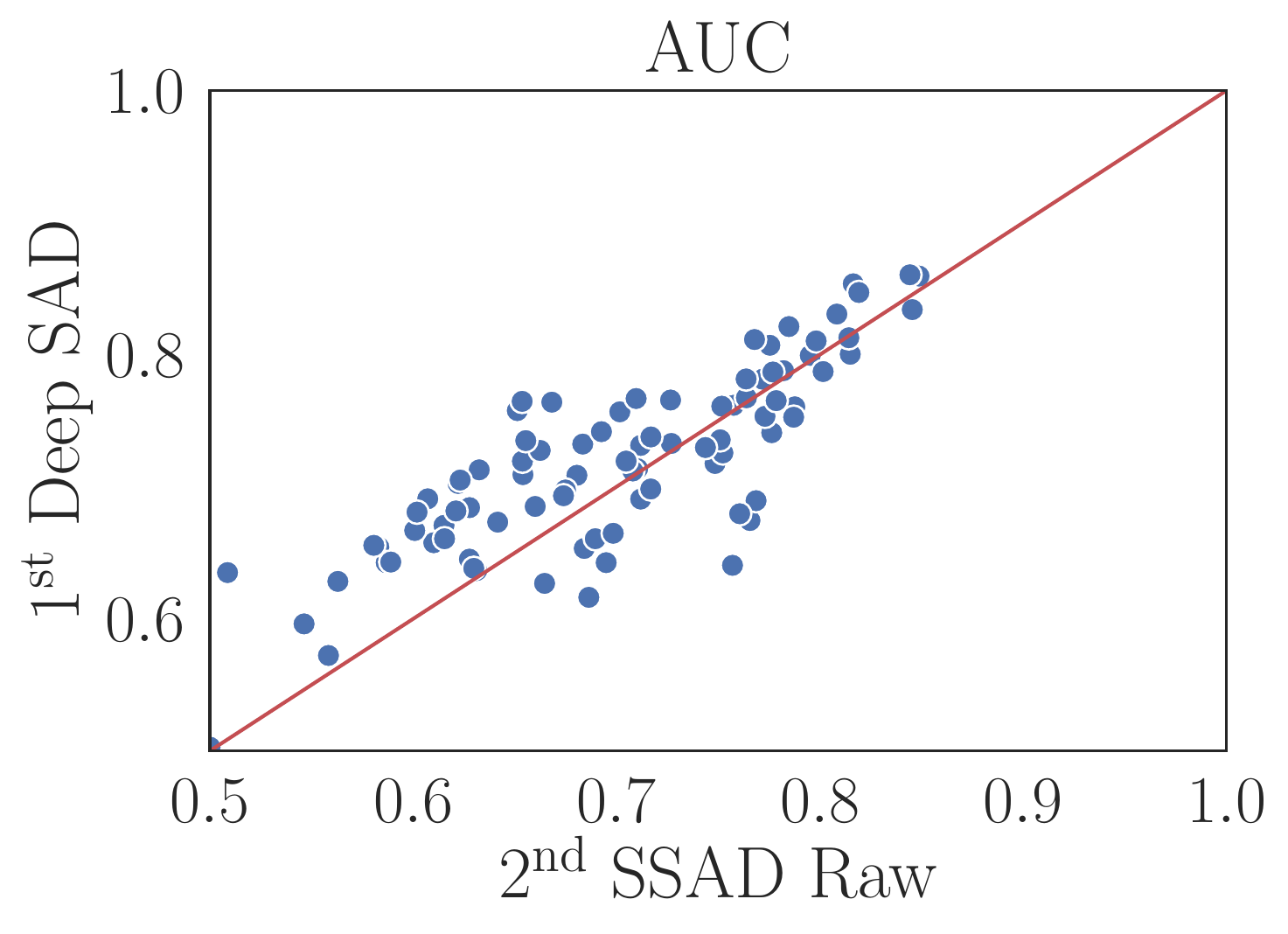}}
\subfigure[$\pol = 0.2$]{\label{fig:pollution_scatter_e}\includegraphics[width=0.329\linewidth]{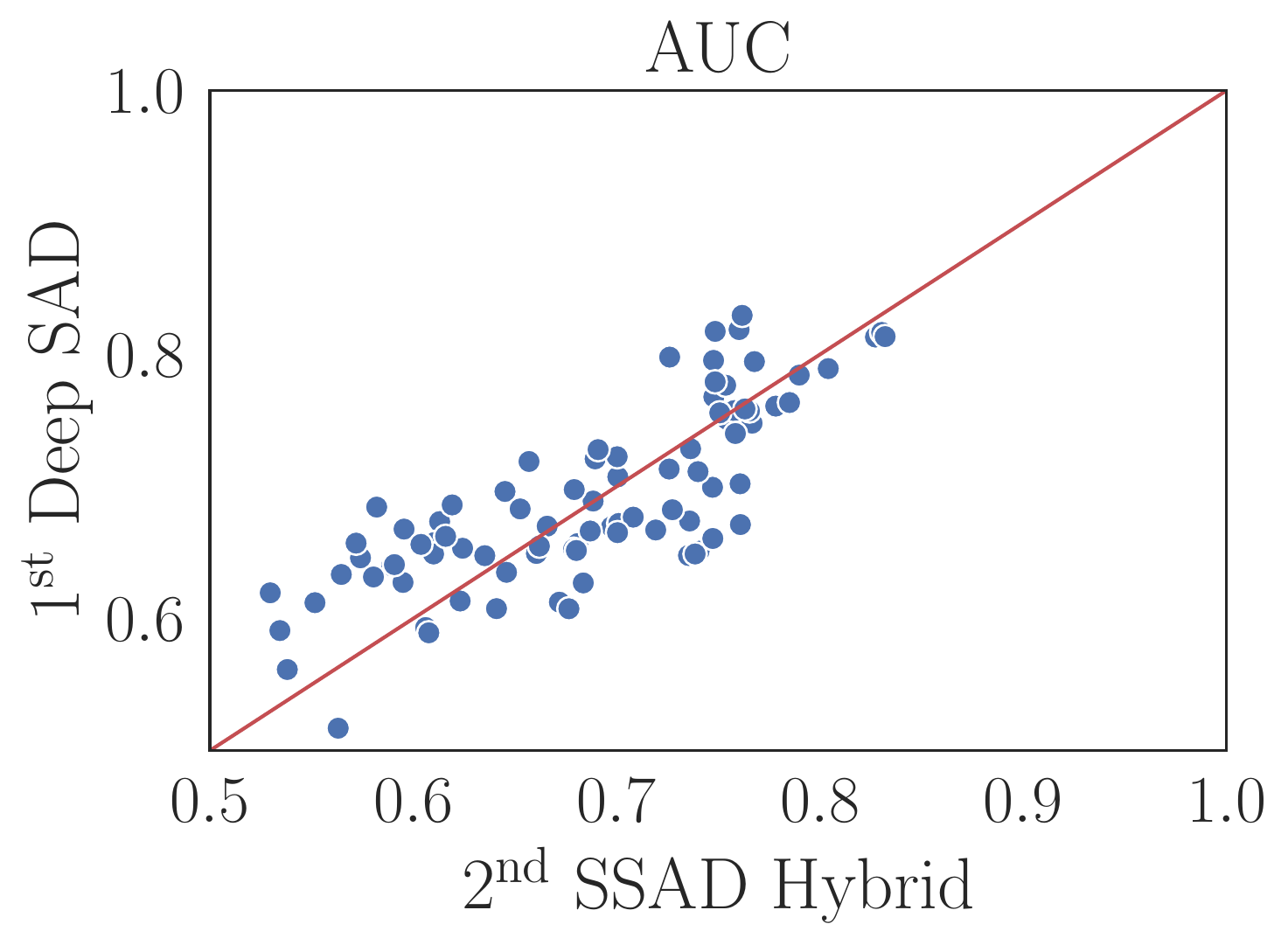}}
\caption{AUC scatterplots of best (1\textsuperscript{st}) vs.~second best (2\textsuperscript{nd}) performing methods in experimental scenario (ii) on CIFAR-10, where we pollute the unlabeled part of the training set with (unknown) anomalies at various ratios $\pol$.}
\label{fig:pollution_scatter}
\end{figure}

\begin{figure}[ht]
\centering
\subfigure[$\klab = 1$]{\label{fig:n_known_scatter_a}\includegraphics[width=0.329\linewidth]{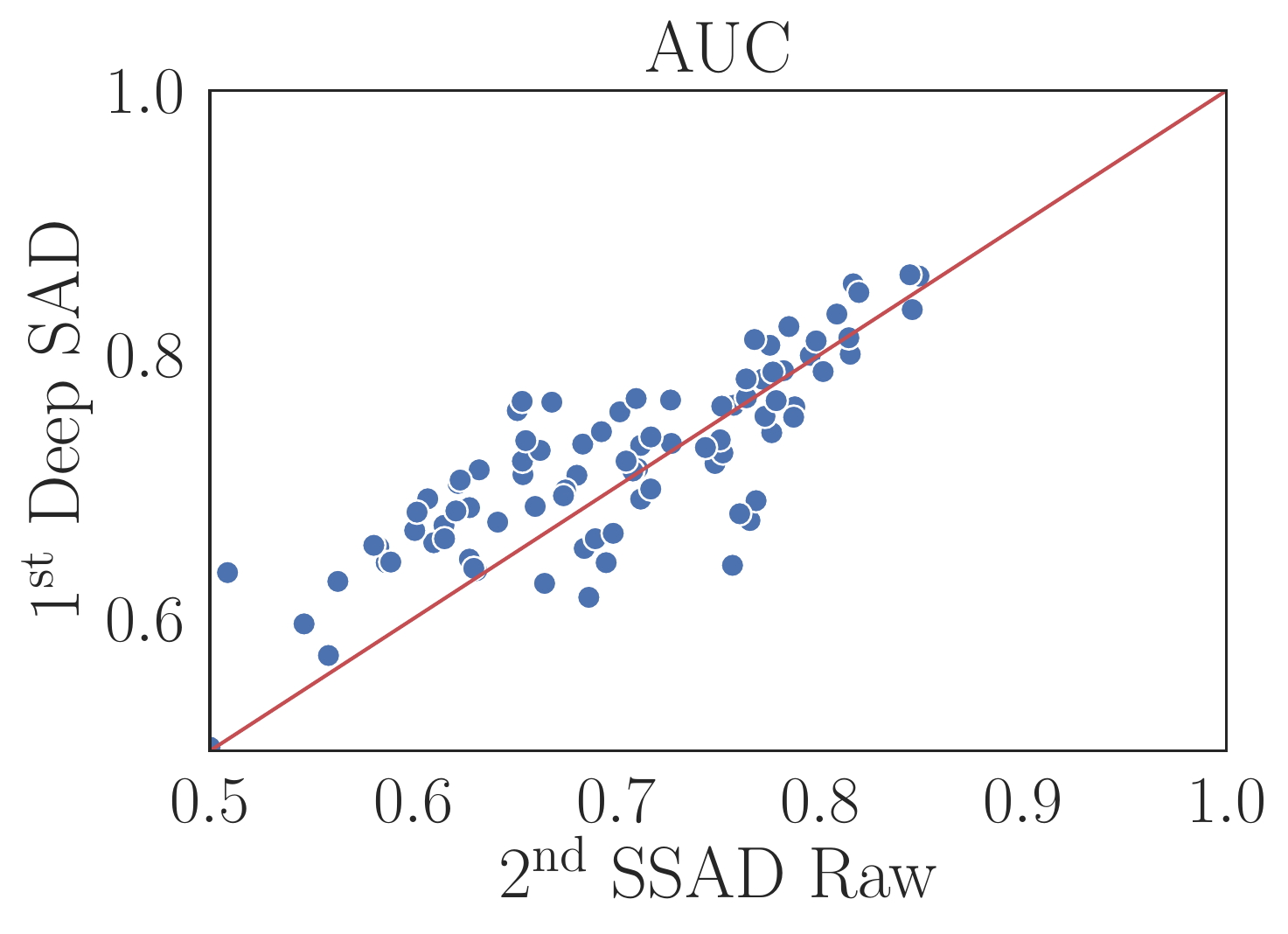}}
\subfigure[$\klab = 2$]{\label{fig:n_known_scatter_b}\includegraphics[width=0.329\linewidth]{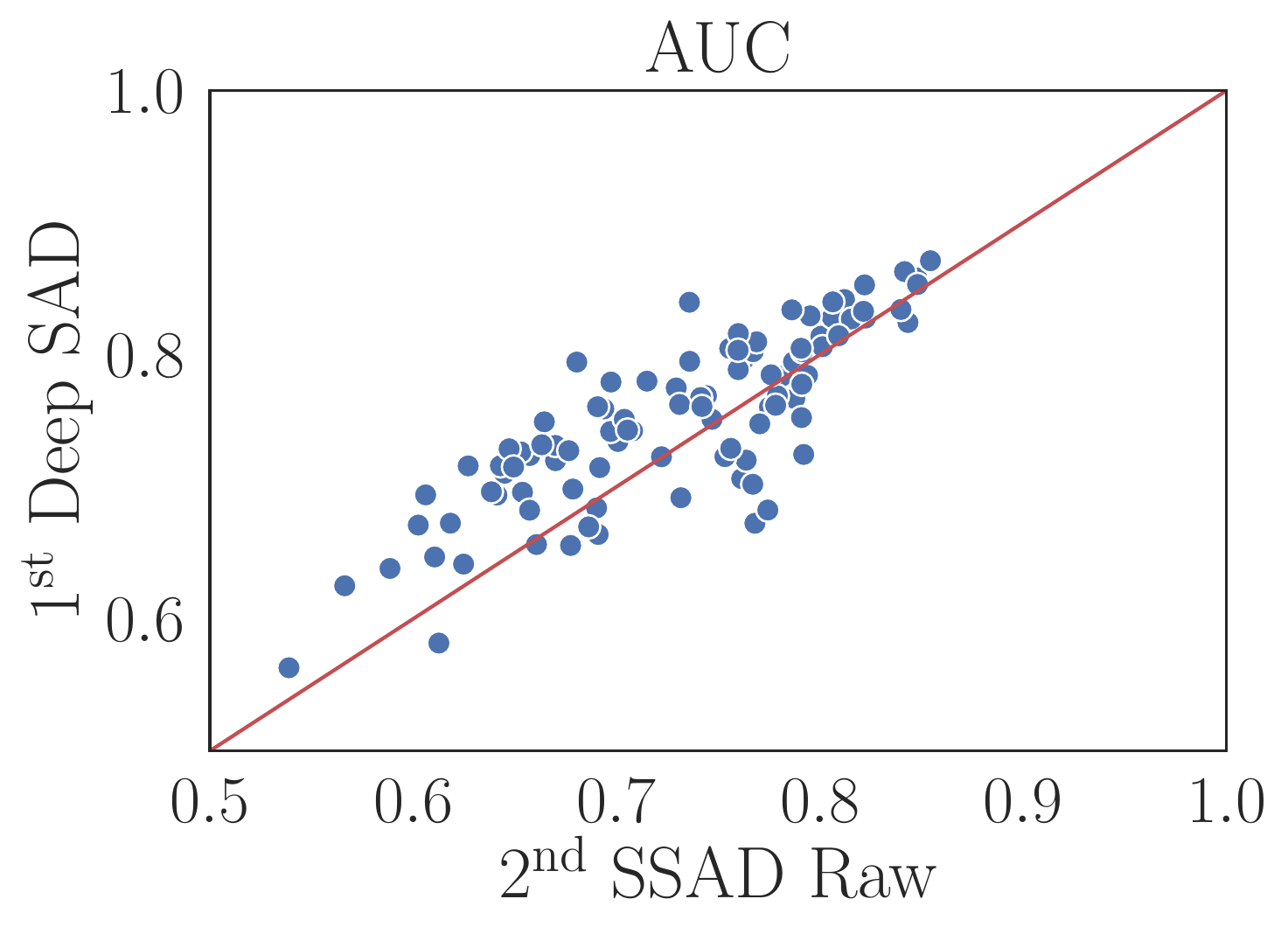}}
\subfigure[$\klab = 3$]{\label{fig:n_known_scatter_c}\includegraphics[width=0.329\linewidth]{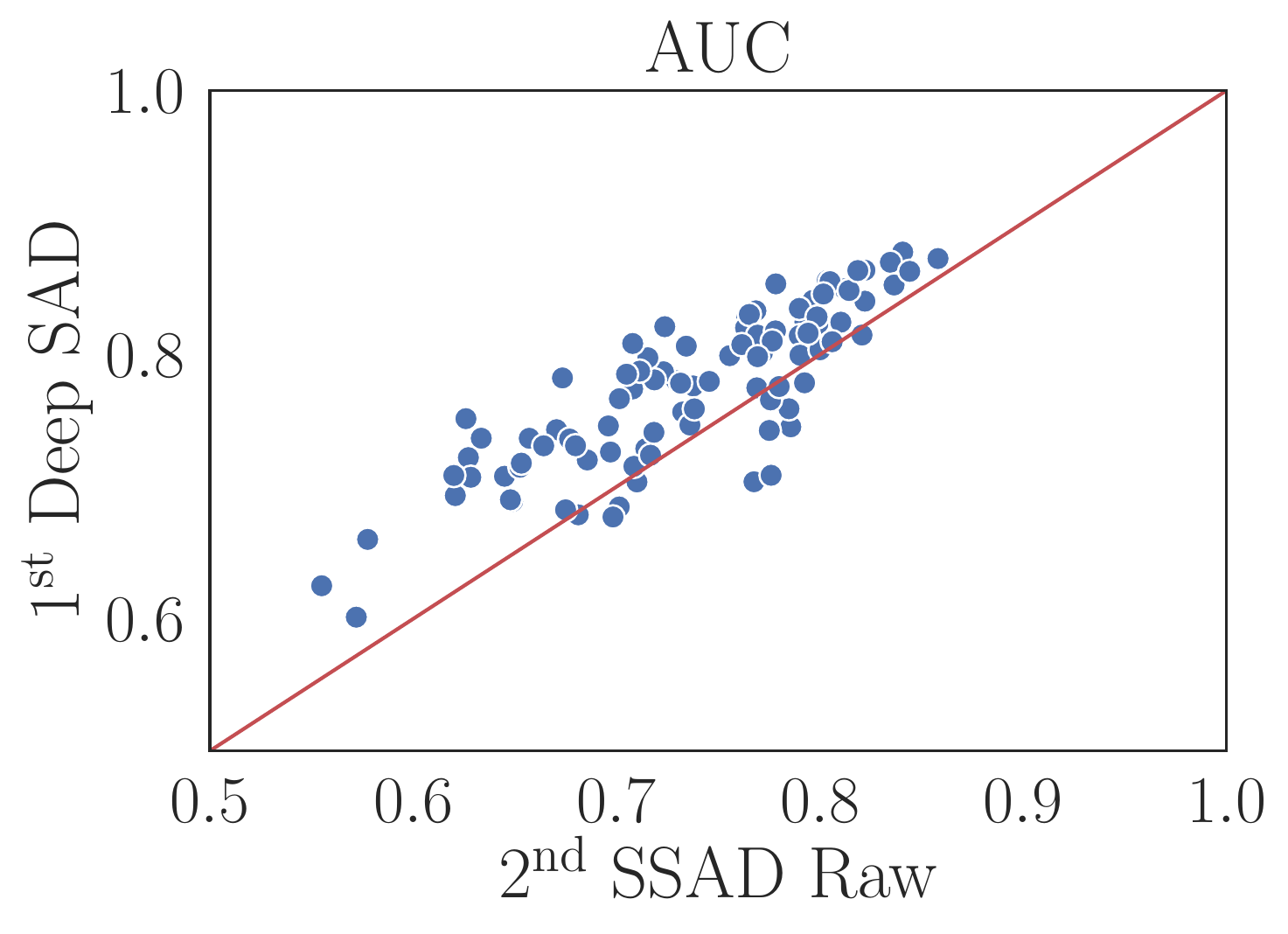}}
\subfigure[$\klab = 5$]{\label{fig:n_known_scatter_d}\includegraphics[width=0.329\linewidth]{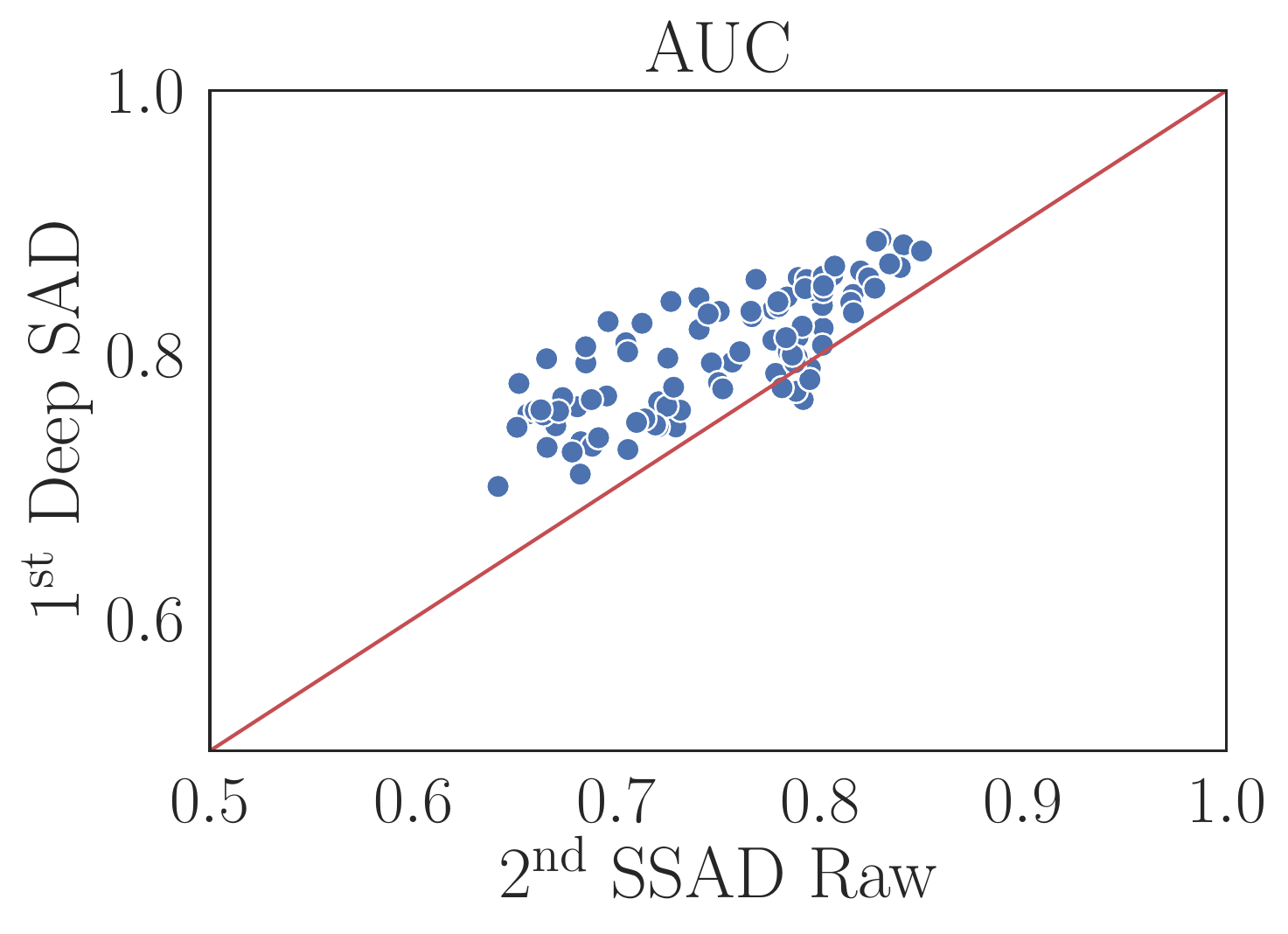}}
\caption{AUC scatterplots of best (1\textsuperscript{st}) vs.~second best (2\textsuperscript{nd}) performing methods in experimental scenario (iii) on CIFAR-10, where we increase the number of anomaly classes $\klab$ included in the labeled training data. }
\label{fig:n_known_scatter}
\end{figure}

We provide AUC scatterplots in Figures \ref{fig:1_known_scatter}--\ref{fig:n_known_scatter} of the best (1\textsuperscript{st}) vs.~second best (2\textsuperscript{nd}) performing methods in the experimental scenarios (i)--(iii) on the most complex CIFAR-10 dataset.
If most points fall above the identity line, this is a very strong indication that the best method indeed significantly outperforms the second best, which often is the case for our Deep SAD method.

%%%%%%%%%%%%%%%%%%%%%%%%%%%%%%%%%%%%%%%%%%%%%%%%%%%%%%%%%%%%%%%%%%%%%%%%%%%%%%%%
%%%%%%%%%%%%%%%%%%%%%%%%%%%%%%%%%%%%%%%%%%%%%%%%%%%%%%%%%%%%%%%%%%%%%%%%%%%%%%%%

%%%%%%%%%%%%%%%%%%%%%%%%%%%%%%%%%%%%%%%%%%%%%%%%%%%%%%%%%%%%%%%%%%%%%%%%%%%%%%%%
%%%%%%%%%%%%%%%%%%%%%%%%%%%%%%%%%%%%%%%%%%%%%%%%%%%%%%%%%%%%%%%%%%%%%%%%%%%%%%%%

%%%%%%%%%%%%%%%%%%%%%%%%%%%%%%%%%%%%%%%%%%%%%%%%%%%%%%%%%%%%%%%%%%%%%%%%%%%%%%%%
%%%%%%%%%%%%%%%%%%%%%%%%%%%%%%%%%%%%%%%%%%%%%%%%%%%%%%%%%%%%%%%%%%%%%%%%%%%%%%%%

\clearpage

\section{Results on Classic Anomaly Detection Benchmark Datasets}
\label{sec:app_exp_classic}

%%%%%%%%%%%%%%%%%%%%%%%%%%%%%%%%%%%%%%%%%%%%%%%%%%%%%%%%%%%%%%%%%%%%%%%%%%%%%%%%
\begin{wraptable}{R}{0.5\textwidth}
    \centering
    \caption{Anomaly detection benchmarks.}
    \label{tab:odds_details}
    \begin{small}
    \begin{tabular}{lrrr}
        \toprule
        \textbf{Dataset}    & $N$       & $D$       & \#outliers (\%)\\
        \midrule
        arrhythmia          & 452       & 274       & 66 (14.6\%) \\
        cardio              & 1,831     & 21        & 176 (9.6\%) \\
        satellite           & 6,435     & 36        & 2,036 (31.6\%) \\
        satimage-2          & 5,803     & 36        & 71 (1.2\%) \\
        shuttle             & 49,097    & 9         & 3,511 (7.2\%) \\
        thyroid             & 3,772     & 6         & 93 (2.5\%) \\
        \bottomrule
    \end{tabular}
    \end{small}
\end{wraptable}
%%%%%%%%%%%%%%%%%%%%%%%%%%%%%%%%%%%%%%%%%%%%%%%%%%%%%%%%%%%%%%%%%%%%%%%%%%%%%%%%
In this experiment, we examine the detection performance on some well-established AD benchmark datasets \citep{rayana2016} listed in Table~\ref{tab:odds_details}.
We do this to evaluate the deep against the shallow approaches also on non-image, tabular datasets that are rarely considered in the deep AD literature.
For the evaluation, we consider random train-to-test set splits of 60:40 while maintaining the original proportion of anomalies in each set.
We then run experiments for 10 seeds with $\gamma_l = 0.01$ and $\gamma_p = 0$, i.e.~1\% of the training set are labeled anomalies and the unlabeled training data is unpolluted.
Since there are no specific different anomaly classes in these datasets, we have $k_l = 1$.
We standardize features to have zero mean and unit variance as the only pre-processing step.

Table~\ref{tab:odds_results} shows the results of the competitive methods.
We observe that the shallow kernel methods seem to perform slightly better on the rather small, low-dimensional benchmarks.
Deep SAD proves competitive though and the small differences might be explained by the strong advantage we grant the shallow methods in the selection of their hyperparameters.
We provide the complete table with the results from all methods in Appendix \ref{sec:app_full_results}

%%%%%%%%%%%%%%%%%%%%%%%%%%%%%%%%%%%%%%%%%%%%%%%%%%%%%%%%%%%%%%%%%%%%%%%%%%%%%%%%
\begin{table}[ht]
\caption{Results on classic AD benchmark datasets in the setting with no pollution $\gamma_p = 0$ and a ratio of labeled anomalies of $\gamma_l = 0.01$ in the training set. We report avg.~AUC with st.~dev.~computed over 10 seeds. A ``$\star$'' indicates a statistically significant ($\alpha=0.05$) difference between 1\textsuperscript{st} and 2\textsuperscript{nd}.}
\label{tab:odds_results}
\begin{center}
\begin{small}
\begin{tabular}{lccccccc}
\toprule
                    & \textbf{OC-SVM}       & \textbf{OC-SVM}       & \textbf{Deep}         & \textbf{SSAD}         & \textbf{SSAD}         & \textbf{Supervised}       & \textbf{Deep}\\
\textbf{Dataset}    & \textbf{Raw}          & \textbf{Hybrid}       & \textbf{SVDD} 	    & \textbf{Raw}          & \textbf{Hybrid}       & \textbf{Classifier}       & \textbf{SAD}\\
\midrule
arrhythmia          & 84.5$\pm$3.9          & 76.7$\pm$6.2          & 74.6$\pm$9.0          & \textbf{86.7$\pm$4.0$^\star$} & 78.3$\pm$5.1          & 39.2$\pm$9.5              & 75.9$\pm$8.7\\
cardio              & 98.5$\pm$0.3          & 82.8$\pm$9.3          & 84.8$\pm$3.6          & \textbf{98.8$\pm$0.3} & 86.3$\pm$5.8          & 83.2$\pm$9.6              & 95.0$\pm$1.6\\
satellite           & 95.1$\pm$0.2          & 68.6$\pm$4.8          & 79.8$\pm$4.1          & \textbf{96.2$\pm$0.3$^\star$} & 86.9$\pm$2.8          & 87.2$\pm$2.1              & 91.5$\pm$1.1\\
satimage-2          & 99.4$\pm$0.8          & 96.7$\pm$2.1          & 98.3$\pm$1.4          & \textbf{99.9$\pm$0.1} & 96.8$\pm$2.1          & \textbf{99.9$\pm$0.1}     & \textbf{99.9$\pm$0.1}\\
shuttle             & 99.4$\pm$0.9          & 94.1$\pm$9.5          & 86.3$\pm$7.5         & \textbf{99.6$\pm$0.5}  & 97.7$\pm$1.0          & 95.1$\pm$8.0              & 98.4$\pm$0.9\\
thyroid             & 98.3$\pm$0.9          & 91.2$\pm$4.0          & 72.0$\pm$9.7          & 97.9$\pm$1.9          & 95.3$\pm$3.1          & 97.8$\pm$2.6              & \textbf{98.6$\pm$0.9}\\
\bottomrule
\end{tabular}
\end{small}
\end{center}
\end{table}
%%%%%%%%%%%%%%%%%%%%%%%%%%%%%%%%%%%%%%%%%%%%%%%%%%%%%%%%%%%%%%%%%%%%%%%%%%%%%%%%

\clearpage

\section{Optimization of Deep SAD}
\label{sec:app_optimization}

Our Deep SAD objective (\ref{eq:deepSAD}) is generally non-convex in the network weights $\mathcal{W}$ which usually is the case in deep learning.
For a computationally efficient optimization, we rely on (mini-batch) SGD to optimize the network weights using backpropagation.
For improved generalization, we add $L^2$ weight decay regularization with hyperparameter $\lambda > 0$ to the objective.
Algorithm~\ref{alg:semi_deepSVDD} summarizes the Deep SAD optimization routine.

{\centering
\begin{minipage}{.9\linewidth}
    \begin{algorithm}[H]
        \caption{Optimization of Deep SAD}
        \label{alg:semi_deepSVDD}
        \begin{algorithmic}[1]
        	\Input
              \Statex Unlabeled data: $\bm{x}_1, \ldots, \bm{x}_n$
              \Statex Labeled data: $(\bm{x}'_{1}, y'_{1}), \ldots, (\bm{x}'_{m}, y'_{m})$
              \Statex Hyperparameters: $\eta, \lambda$
              \Statex SGD learning rate: $\varepsilon$
            \Output
              \Statex Trained model: $\mathcal{W}^*$
         	\Statex
            
            \State \textbf{Initialize:}
            \Statex \hspace*{\algorithmicindent} Neural network weights: $\mathcal{W}$
            \Statex \hspace*{\algorithmicindent} Hypersphere center: $\bm{c}$
            
            \For{each epoch}
            	\For{each mini-batch}
                	\State Draw mini-batch $\mathcal{B}$
           			\State $\mathcal{W} \leftarrow \mathcal{W} - \varepsilon \cdot \nabla_{\mathcal{W}} J(\mathcal{W}; \mathcal{B})$ 
            	\EndFor
            \EndFor
        \end{algorithmic}
    \end{algorithm}
\end{minipage}
\vspace{1.5em}
\par
}

Using SGD allows Deep SAD to scale with large datasets as the computational complexity scales linearly in the number of training batches and computations in each batch can be parallelized (e.g., by training on GPUs).
Moreover, Deep SAD has low memory complexity as a trained model is fully characterized by the final network parameters $\mathcal{W}^*$ and no data must be saved or referenced for prediction.
Instead, the prediction only requires a forward pass on the network which usually is just a concatenation of simple functions.
This enables fast predictions for Deep SAD.

\textbf{Initialization of the network weights $\mathcal{W}$ \;} We establish an autoencoder pre-training routine for initialization.
That is, we first train an autoencoder that has an encoder with the same architecture as network $\phi$ on the reconstruction loss (mean squared error or cross-entropy).
After training, we then initialize $\mathcal{W}$ with the converged parameters of the encoder.
Note that this is in line with the Infomax principle (\ref{eq:infomax}) for unsupervised representation learning \citep{vincent2008}.

\textbf{Initialization of the center $\bm{c}$ \;} After initializing the network weights $\mathcal{W}$, we fix the hypersphere center $\bm{c}$ as the mean of the network representations that we obtain from an initial forward pass on the data (excluding labeled anomalies).
We found SGD convergence to be smoother and faster by fixing center $\bm{c}$ in the neighborhood of the initial data representations as also observed by \citet{ruff2018}.
If sufficiently many labeled normal examples are available, using only those examples for a mean initialization would be another strategy to minimize possible distortions from polluted unlabeled training data.
Adding center $\bm{c}$ as a free optimization variable would allow a trivial ``hypersphere collapse'' solution for the fully unlabeled setting, i.e.~for unsupervised Deep SVDD.

\textbf{Preventing a hypersphere collapse \;} A ``hypersphere collapse'' describes the trivial solution that neural network $\phi$ converges to the constant function $\phi \equiv \bm{c}$, i.e.~the hypersphere collapses to a single point.
\citet{ruff2018} demonstrate theoretical network properties that prevent such a collapse which we adopt for Deep SAD.
Most importantly, network $\phi$ must have no bias terms and no bounded activation functions.
We refer to \citet{ruff2018} for further details.
If there are sufficiently many labeled anomalies available for training, however, hypersphere collapse is not a problem for Deep SAD due to the opposing labeled and unlabeled objectives.

\section{Network Architectures}
\label{sec:app_architectures}
We employ LeNet-type convolutional neural networks (CNNs) on MNIST, Fashion-MNIST, and CIFAR-10, where each convolutional module consists of a convolutional layer followed by leaky ReLU activations with leakiness $\alpha=0.1$ and $(2{\times}2)$-max-pooling.
On MNIST, we employ a CNN with two modules, $8{\times}(5{\times}5)$-filters followed by $4{\times}(5{\times}5)$-filters, and a final dense layer of $32$ units.
On Fashion-MNIST, we employ a CNN also with two modules, $16{\times}(5{\times}5)$-filters and $32{\times}(5{\times}5)$-filters, followed by two dense layers of $64$ and $32$ units respectively.
On CIFAR-10, we employ a CNN with three modules, $32{\times}(5{\times}5)$-filters, $64{\times}(5{\times}5)$-filters, and $128{\times}(5{\times}5)$-filters, followed by a final dense layer of $128$ units.

On the classic AD benchmark datasets, we employ standard MLP feed-forward architectures.
On arrhythmia, a 3-layer MLP with $128$-$64$-$32$ units.
On cardio, satellite, satimage-2, and shuttle a 3-layer MLP with $32$-$16$-$8$ units.
On thyroid a 3-layer MLP with $32$-$16$-$4$ units.

For the (convolutional) autoencoders, we always employ the above architectures for the encoder networks and then construct the decoder networks symmetrically, where we replace max-pooling with simple upsampling and convolutions with deconvolutions.

\section{Details on Competing Methods}
\label{sec:app_competitors}

\textbf{OC-SVM/SVDD \;} The OC-SVM and SVDD are equivalent for the Gaussian/RBF kernel we employ.
As mentioned in the main paper, we deliberately grant the OC-SVM/SVDD an unfair advantage by selecting its hyperparameters to maximize AUC on a subset (10\%) of the test set to establish a strong baseline.
To do this, we consider the RBF scale parameter $\gamma \in \{2^{-7}, 2^{-6}, \ldots 2^{2}\}$ and select the best performing one.
Moreover, we always repeat this over $\nu$-parameter $\nu \in \{0.01, 0.05, 0.1, 0.2, 0.5 \}$ and then report the best final result.

\textbf{Isolation Forest (IF) \;} We set the number of trees to $t = 100$ and the sub-sampling size to $\psi = 256$, as recommended in the original work \citep{liu2008}. 

\textbf{Kernel Density Estimator (KDE) \;} We select the bandwidth $h$ of the Gaussian kernel from $h \in \{2^{0.5}, 2^1, \ldots ,2^5 \}$ via 5-fold cross-validation using the log-likelihood score following \citep{ruff2018}.

\textbf{SSAD \;} We also deliberately grant the state-of-the-art semi-supervised AD kernel method SSAD the unfair advantage of selecting its hyperparameters optimally to maximize AUC on a subset (10\%) of the test set.
To do this, we again select the scale parameter $\gamma$ of the RBF kernel we use from $\gamma \in \{2^{-7}, 2^{-6}, \ldots 2^{2}\}$ and select the best performing one.
Otherwise we set the hyperparameters as recommend by the original authors to $\kappa = 1$, $\kappa = 1$, $\eta_u = 1$, and $\eta_l = 1$ \citep{gornitz2013}.

\textbf{(Convolutional) Autoencoder ((C)AE) \;} To create the (convolutional) autoencoders, we symmetrically construct the decoders w.r.t.~the architectures reported in Appenidx \ref{sec:app_architectures}, which make up the encoder parts of the autoencoders.
Here, we replace max-pooling with simple upsampling and convolutions with deconvolutions.
We train the autoencoders on the MSE reconstruction loss that also serves as the anomaly score.

\textbf{Hybrid Variants \;} To establish hybrid methods, we apply the OC-SVM, IF, KDE, and SSAD as outlined above to the resulting bottleneck representations given by the respective converged autoencoders.

\textbf{Unsupervised Deep SVDD \;} We consider both variants, Soft-Boundary Deep SVDD and One-Class Deep SVDD as unsupervised baselines and always report the better performance as the unsupervised result.
For Soft-Boundary Deep SVDD, we optimally solve for the radius $R$ on every mini-batch and run experiments for $\nu \in \{0.01, 0.1 \}$.
We set the weight decay hyperparameter to $\lambda = 10^{-6}$.
For Deep SVDD, we always remove all the bias terms from a network to prevent a hypersphere collapse as recommended by the authors in the original work \citep{ruff2018}.

\textbf{Deep SAD \;} We set $\lambda = 10^{-6}$ and equally weight the unlabeled and labeled examples by setting $\eta = 1$ if not reported otherwise.

\textbf{SS-DGM \;} We consider both the M2 and M1+M2 model and always report the better performing result.
Otherwise we follow the settings as recommended in the original work \citep{kingma2014b}.

Note that we use the latent \emph{class probability estimate} (normal vs.~anomalous) of semi-supervised DGM as a natural choice for the anomaly score, and \emph{not} the reconstruction error as used for unsupervised autoencoding models such as the (convolutional) autoencoder we consider.
Such deep semi-supervised models designed for classification as the downstream task have no notion of out-of-distribution and again implicitly make the cluster assumption \citep{zhu2005,chapelle2009} we refer to.
Thus, semi-supervised DGM also suffers from overfitting to previously seen anomalies at training similar to the supervised model which explains its bad AD performance.

\textbf{Supervised Deep Binary Classifier \;} To interpret AD as a binary classification problem, we rely on the typical assumption that most of the unlabeled training data is normal by assigning $y={+}1$ to all unlabeled examples.
Already labeled normal examples and labeled anomalies retain their assigned labels of $\tilde{y} = {+}1$ and $\tilde{y} = {-}1$ respectively.
We train the supervised classifier on the binary cross-entropy loss.
Note that in scenario (i), in particular, the supervised classifier has perfect, unpolluted label information but still fails to generalize as there are novel anomaly classes at testing.

\textbf{SGD Optimization Details for Deep Methods \;} We use the Adam optimizer with recommended default hyperparameters \citep{kingma2014} and apply Batch Normalization \citep{ioffe2015} in SGD optimization.
For all deep approaches and on all datasets, we employ a two-phase (``searching'' and ``fine-tuning'') learning rate schedule.
In the searching phase we first train with a learning rate $\varepsilon = 10^{-4}$ for $50$ epochs. 
In the fine-tuning phase we train with $\varepsilon = 10^{-5}$ for another $100$ epochs.
We always use a batch size of 200.
For the autoencoder, SS-DGM, and the supervised classifier, we initialize the network with uniform Glorot weights \citep{glorot2010}.
For Deep SVDD and Deep SAD, we establish an unsupervised pre-training routine via autoencoder as explained in Appendix \ref{sec:app_optimization}, where we set the network $\phi$ to be the encoder of the autoencoder that we train beforehand.

\section{Complete Tables of Experimental Results}
\label{sec:app_full_results}
The following Tables \ref{tab:1_known}--\ref{tab:odds_results_full} list the complete experimental results of all the methods in all our experiments.

\clearpage

%%%%%%%%%%%%%%%%%%%%%%%%%%%%%%%%%%%%%%%%%%%%%%%%%%%%%%%%%%%%%%%%%%%%%%%%%%%%%%%%
\afterpage{%
    \clearpage%
    \begin{landscape}% Landscape page
        \centering % Center table
        \captionsetup{type=table}
        \captionof{table}{Complete results of experimental scenario (i), where we increase the ratio of labeled anomalies $\lab$ in the training set. We report the avg.~AUC with st.~dev.~computed over 90 experiments at various ratios $\lab$.}%
        \label{tab:1_known}
        \vspace{1em}
        \begin{scriptsize}
        %%%%%%%%%%%%%%%%%%%%%%%%%%%%%%%%%%%%%%%%
\begin{tabular}{llccccccccccccc}
\toprule
                    &       & \textbf{OC-SVM}   & \textbf{OC-SVM}   & \textbf{IF}   & \textbf{IF}       & \textbf{KDE}      & \textbf{KDE}      &                   & \textbf{Deep}     & \textbf{SSAD} & \textbf{SSAD}     &                   &   \textbf{Deep}   & \textbf{Supervised} \\
\textbf{Data}	    &$\lab$ & \textbf{Raw}      & \textbf{Hybrid}   & \textbf{Raw}  & \textbf{Hybrid}   & \textbf{Raw}      & \textbf{Hybrid}   & \textbf{CAE}      & \textbf{SVDD}     & \textbf{Raw}  & \textbf{Hybrid}   & \textbf{SS-DGM}   &   \textbf{SAD}   & \textbf{Classifier} \\
\midrule
MNIST 				& .00   & 96.0$\pm$2.9      & 96.3$\pm$2.5      & 85.4$\pm$8.7  & 90.5$\pm$5.3      & 95.0$\pm$3.3      & 87.8$\pm$5.6      & 92.9$\pm$5.7      & 92.8$\pm$4.9      & 96.0$\pm$2.9  & 96.3$\pm$2.5      &                   & 92.8$\pm$4.9 &      \\
	 				& .01   &                   &                   &               &                   &                   &                   &                   &                   & 96.6$\pm$2.4  & 96.8$\pm$2.3      & 89.9$\pm$9.2      & 96.4$\pm$2.7 & 92.8$\pm$5.5                \\
					& .05   &                   &                   &               &                   &                   &                   &                   &                   & 93.3$\pm$3.6  & 97.4$\pm$2.0      & 92.2$\pm$5.6      & 96.7$\pm$2.4 & 94.5$\pm$4.6                \\
	 				& .10   &                   &                   &               &                   &                   &                   &                   &                   & 90.7$\pm$4.4  & 97.6$\pm$1.7      & 91.6$\pm$5.5      & 96.9$\pm$2.3 & 95.0$\pm$4.7                \\
	 				& .20   &                   &                   &               &                   &                   &                   &                   &                   & 87.2$\pm$5.6  & 97.8$\pm$1.5      & 91.2$\pm$5.6      & 96.9$\pm$2.4 & 95.6$\pm$4.4                \\
\midrule
F-MNIST             & .00   & 92.8$\pm$4.7      & 91.2$\pm$4.7      & 91.6$\pm$5.5  & 82.5$\pm$8.1      & 92.0$\pm$4.9      & 69.7$\pm$14.4     & 90.2$\pm$5.8      & 89.2$\pm$6.2      & 92.8$\pm$4.7  & 91.2$\pm$4.7      &                   & 89.2$\pm$6.2 &                        \\
	 				& .01   &                   &                   &               &                   &                   &                   &                   &                   & 92.1$\pm$5.0  & 89.4$\pm$6.0      & 65.1$\pm$16.3     & 90.0$\pm$6.4 & 74.4$\pm$13.6          \\
					& .05   &                   &                   &               &                   &                   &                   &                   &                   & 88.3$\pm$6.2  & 90.5$\pm$5.9      & 71.4$\pm$12.7     & 90.5$\pm$6.5 & 76.8$\pm$13.2          \\
	 				& .10   &                   &                   &               &                   &                   &                   &                   &                   & 85.5$\pm$7.1  & 91.0$\pm$5.6      & 72.9$\pm$12.2     & 91.3$\pm$6.0 & 79.0$\pm$12.3          \\
	 				& .20   &                   &                   &               &                   &                   &                   &                   &                   & 82.0$\pm$8.0  & 89.7$\pm$6.6      & 74.7$\pm$13.5     & 91.0$\pm$5.5 & 81.4$\pm$12.0          \\
\midrule
CIFAR-10	 		& .00   & 62.0$\pm$10.6     & 63.8$\pm$9.0      & 60.0$\pm$10.0 & 59.9$\pm$6.7      & 59.9$\pm$11.7     & 56.1$\pm$10.2     & 56.2$\pm$13.2     & 60.9$\pm$9.4      & 62.0$\pm$10.6 & 63.8$\pm$9.0      &                   & 60.9$\pm$9.4 &                        \\
	 				& .01   &                   &                   &               &                   &                   &                   &                   &                   & 73.0$\pm$8.0  & 70.5$\pm$8.3      & 49.7$\pm$1.7      & 72.6$\pm$7.4 & 55.6$\pm$5.0           \\
					& .05   &                   &                   &               &                   &                   &                   &                   &                   & 71.5$\pm$8.1  & 73.3$\pm$8.4      & 50.8$\pm$4.7      & 77.9$\pm$7.2 & 63.5$\pm$8.0           \\
	 				& .10   &                   &                   &               &                   &                   &                   &                   &                   & 70.1$\pm$8.1  & 74.0$\pm$8.1      & 52.0$\pm$5.5      & 79.8$\pm$7.1 & 67.7$\pm$9.6           \\
	 				& .20   &                   &                   &               &                   &                   &                   &                   &                   & 67.4$\pm$8.8  & 74.5$\pm$8.0      & 53.2$\pm$6.7      & 81.9$\pm$7.0 & 80.5$\pm$5.9           \\
\bottomrule
\end{tabular}
        %%%%%%%%%%%%%%%%%%%%%%%%%%%%%%%%%%%%%%%%
        \end{scriptsize}
    \end{landscape}
    \clearpage% Flush page
}
%%%%%%%%%%%%%%%%%%%%%%%%%%%%%%%%%%%%%%%%%%%%%%%%%%%%%%%%%%%%%%%%%%%%%%%%%%%%%%%%

%%%%%%%%%%%%%%%%%%%%%%%%%%%%%%%%%%%%%%%%%%%%%%%%%%%%%%%%%%%%%%%%%%%%%%%%%%%%%%%%
\afterpage{%
    \clearpage%
    \begin{landscape}% Landscape page
        \centering % Center table
        \captionsetup{type=table}
        \captionof{table}{Complete results of experimental scenario (ii), where we pollute the unlabeled part of the training set with (unknown) anomalies. We report the avg.~AUC with st.~dev.~computed over 90 experiments at various ratios $\pol$.}%
        \label{tab:pollution}
        \vspace{1em}
        \begin{scriptsize}
        %%%%%%%%%%%%%%%%%%%%%%%%%%%%%%%%%%%%%%%%
        \begin{tabular}{llccccccccccccc}
        \toprule
                            &       & \textbf{OC-SVM}   & \textbf{OC-SVM}   & \textbf{IF}           & \textbf{IF}           & \textbf{KDE}          & \textbf{KDE}      &                   & \textbf{Deep}     & \textbf{SSAD}     & \textbf{SSAD}     &                   & \textbf{Deep}     & \textbf{Supervised} \\
        \textbf{Data}	    & $\pol$& \textbf{Raw}      & \textbf{Hybrid}   & \textbf{Raw}          & \textbf{Hybrid}       & \textbf{Raw}          & \textbf{Hybrid}   & \textbf{CAE}      & \textbf{SVDD}     & \textbf{Raw}      & \textbf{Hybrid}   & \textbf{SS-DGM}   & \textbf{SAD}      & \textbf{Classifier} \\
        \midrule
        MNIST 				& .00 	& 96.0$\pm$2.9      & 96.3$\pm$2.5      & 85.4$\pm$8.7          & 90.5$\pm$5.3          & 95.0$\pm$3.3          & 87.8$\pm$5.6      & 92.9$\pm$5.7      & 92.8$\pm$4.9      & 97.9$\pm$1.8      & 97.4$\pm$2.0      & 92.2$\pm$5.6      & 96.7$\pm$2.4      & 94.5$\pm$4.6\\
        	 				& .01 	& 94.3$\pm$3.9      & 95.6$\pm$2.5      & 85.2$\pm$8.8          & 90.6$\pm$5.0          & 91.2$\pm$4.9          & 87.9$\pm$5.3      & 91.3$\pm$6.1      & 92.1$\pm$5.1      & 96.6$\pm$2.4      & 95.2$\pm$2.3      & 92.0$\pm$6.0      & 95.5$\pm$3.3      & 91.5$\pm$5.9\\
        					& .05 	& 91.4$\pm$5.2      & 93.8$\pm$3.9      & 83.9$\pm$9.2          & 89.7$\pm$6.0          & 85.5$\pm$7.1          & 87.3$\pm$7.0      & 87.2$\pm$7.1      & 89.4$\pm$5.8      & 93.4$\pm$3.4      & 89.5$\pm$3.9      & 91.0$\pm$6.9      & 93.5$\pm$4.1      & 86.7$\pm$7.4\\
        	 				& .10 	& 88.8$\pm$6.0      & 91.4$\pm$5.1      & 82.3$\pm$9.5          & 88.2$\pm$6.5          & 82.1$\pm$8.5          & 85.9$\pm$6.6      & 83.7$\pm$8.4      & 86.5$\pm$6.8      & 90.7$\pm$4.4      & 86.0$\pm$4.6      & 89.7$\pm$7.5      & 91.2$\pm$4.9      & 83.6$\pm$8.2\\
        	 				& .20 	& 84.1$\pm$7.6      & 85.9$\pm$7.6      & 78.7$\pm$10.5         & 85.3$\pm$7.9          & 77.4$\pm$10.9         & 82.6$\pm$8.6      & 78.6$\pm$10.3     & 81.5$\pm$8.4      & 87.4$\pm$5.6      & 82.1$\pm$5.4      & 87.4$\pm$8.6      & 86.6$\pm$6.6      & 79.7$\pm$9.4\\
        \midrule
        F-MNIST             & .00	& 92.8$\pm$4.7      & 91.2$\pm$4.7      & 91.6$\pm$5.5          & 82.5$\pm$8.1          & 92.0$\pm$4.9          & 69.7$\pm$14.4     & 90.2$\pm$5.8      & 89.2$\pm$6.2      & 94.0$\pm$4.4      & 90.5$\pm$5.9      & 71.4$\pm$12.7     & 90.5$\pm$6.5      & 76.8$\pm$13.2\\
        	 				& .01 	& 91.7$\pm$5.0      & 91.5$\pm$4.6      & 91.5$\pm$5.5          & 84.9$\pm$7.2          & 89.4$\pm$6.3          & 73.9$\pm$12.4     & 87.1$\pm$7.3      & 86.3$\pm$6.3      & 92.2$\pm$4.9      & 87.8$\pm$6.1      & 71.2$\pm$14.3     & 87.2$\pm$7.1      & 67.3$\pm$8.1\\
        					& .05 	& 90.7$\pm$5.5      & 90.7$\pm$4.9      & 90.9$\pm$5.9          & 85.5$\pm$7.2          & 85.2$\pm$9.1          & 75.4$\pm$12.9     & 81.6$\pm$9.6      & 80.6$\pm$7.1      & 88.3$\pm$6.2      & 82.7$\pm$7.8      & 71.9$\pm$14.3     & 81.5$\pm$8.5      & 59.8$\pm$4.6\\
        	 				& .10 	& 89.5$\pm$6.1      & 89.3$\pm$6.2      & 90.2$\pm$6.3          & 85.5$\pm$7.7          & 81.8$\pm$11.2         & 77.8$\pm$12.0     & 77.4$\pm$11.1     & 76.2$\pm$7.3      & 85.6$\pm$7.0      & 79.8$\pm$9.0      & 72.5$\pm$15.5     & 78.2$\pm$9.1      & 56.7$\pm$4.1\\
        	 				& .20 	& 86.3$\pm$7.7      & 88.1$\pm$6.9      & 88.4$\pm$7.6          & 86.3$\pm$7.4          & 77.4$\pm$13.6         & 82.1$\pm$9.8      & 72.5$\pm$12.6     & 69.3$\pm$6.3      & 81.9$\pm$8.1      & 74.3$\pm$10.6     & 70.8$\pm$16.0     & 74.8$\pm$9.4      & 53.9$\pm$2.9\\
        \midrule
        CIFAR-10	 		& .00 	& 62.0$\pm$10.6     & 63.8$\pm$9.0      & 60.0$\pm$10.0         & 59.9$\pm$6.7          & 59.9$\pm$11.7         & 56.1$\pm$10.2     & 56.2$\pm$13.2     & 60.9$\pm$9.4      & 73.8$\pm$7.6      & 73.3$\pm$8.4      & 50.8$\pm$4.7      & 77.9$\pm$7.2      & 63.5$\pm$8.0\\
        	 				& .01 	& 61.9$\pm$10.6     & 63.8$\pm$9.3      & 59.9$\pm$10.1         & 59.9$\pm$6.7          & 59.2$\pm$12.3         & 56.3$\pm$10.4     & 56.2$\pm$13.1     & 60.5$\pm$9.4      & 73.0$\pm$8.0      & 72.8$\pm$8.1      & 51.1$\pm$4.7      & 76.5$\pm$7.2      & 62.9$\pm$7.3\\
        					& .05 	& 61.4$\pm$10.7     & 62.6$\pm$9.2      & 59.6$\pm$10.1         & 59.6$\pm$6.4          & 58.1$\pm$12.9         & 55.6$\pm$10.5     & 55.7$\pm$13.3     & 59.6$\pm$9.8      & 71.5$\pm$8.2      & 71.0$\pm$8.4      & 50.1$\pm$2.9      & 74.0$\pm$6.9      & 62.2$\pm$8.2\\
        	 				& .10 	& 60.8$\pm$10.7     & 62.9$\pm$8.2      & 58.8$\pm$10.1         & 59.1$\pm$6.6          & 57.3$\pm$13.5         & 54.9$\pm$11.1     & 55.4$\pm$13.3     & 58.6$\pm$10.0     & 69.8$\pm$8.4      & 69.3$\pm$8.5      & 50.5$\pm$3.6      & 71.8$\pm$7.0      & 60.6$\pm$8.3\\
        	 				& .20 	& 60.3$\pm$10.3     & 61.9$\pm$8.1      & 57.9$\pm$10.1         & 58.3$\pm$6.2          & 56.2$\pm$13.9         & 54.2$\pm$11.1     & 54.6$\pm$13.3     & 57.0$\pm$10.6     & 67.8$\pm$8.6      & 67.9$\pm$8.1      & 50.1$\pm$1.7      & 68.5$\pm$7.1      & 58.5$\pm$6.7\\
        \bottomrule
        \end{tabular}
        %%%%%%%%%%%%%%%%%%%%%%%%%%%%%%%%%%%%%%%%
        \end{scriptsize}
    \end{landscape}
    \clearpage% Flush page
}
%%%%%%%%%%%%%%%%%%%%%%%%%%%%%%%%%%%%%%%%%%%%%%%%%%%%%%%%%%%%%%%%%%%%%%%%%%%%%%%%

%%%%%%%%%%%%%%%%%%%%%%%%%%%%%%%%%%%%%%%%%%%%%%%%%%%%%%%%%%%%%%%%%%%%%%%%%%%%%%%%
\afterpage{%
    \clearpage%
    \begin{landscape}% Landscape page
        \centering % Center table
        \captionsetup{type=table}
        \captionof{table}{Complete results of experimental scenario (iii), where we increase the number of anomaly classes $\klab$ included in the labeled training data. We report the avg.~AUC with st.~dev.~computed over 100 experiments at various numbers $\klab$.}%
        \label{tab:n_known}
        \vspace{1em}
        \begin{scriptsize}
        %%%%%%%%%%%%%%%%%%%%%%%%%%%%%%%%%%%%%%%%
        \begin{tabular}{llccccccccccccc}
        \toprule
                            &       & \textbf{OC-SVM}   & \textbf{OC-SVM}   & \textbf{IF}   & \textbf{IF}       & \textbf{KDE}  & \textbf{KDE}      &               & \textbf{Deep}     & \textbf{SSAD}     & \textbf{SSAD}     &                   & \textbf{Deep}     & \textbf{Supervised} \\
        \textbf{Data}	    &$\klab$& \textbf{Raw}      & \textbf{Hybrid}   & \textbf{Raw}  & \textbf{Hybrid}   & \textbf{Raw}  & \textbf{Hybrid}   & \textbf{CAE}  & \textbf{SVDD}     & \textbf{Raw}      & \textbf{Hybrid}   & \textbf{SS-DGM}   & \textbf{SAD}      & \textbf{Classifier} \\
        \midrule
        MNIST 				& 0     & 88.8$\pm$6.0      & 91.4$\pm$5.1      & 82.3$\pm$9.5  & 88.2$\pm$6.5      & 82.1$\pm$8.5  & 85.9$\pm$6.6      & 83.7$\pm$8.4  & 86.5$\pm$6.8      & 88.8$\pm$6.0      & 91.4$\pm$5.1      &                   & 86.5$\pm$6.8      & \\
        	 				& 1 	&                   &                   &               &                   &               &                   &               &                   & 90.7$\pm$4.4      & 86.0$\pm$4.6      & 89.7$\pm$7.5      & 91.2$\pm$4.9      & 83.6$\pm$8.2\\
        					& 2 	&                   &                   &               &                   &               &                   &               &                   & 92.5$\pm$3.6      & 87.7$\pm$3.8      & 92.8$\pm$5.3      & 92.0$\pm$3.6      & 90.3$\pm$4.6\\
        	 				& 3 	&                   &                   &               &                   &               &                   &               &                   & 93.9$\pm$3.3      & 89.8$\pm$3.3      & 94.9$\pm$4.2      & 94.7$\pm$2.8      & 93.9$\pm$2.8\\
        	 				& 5 	&                   &                   &               &                   &               &                   &               &                   & 95.5$\pm$2.5      & 91.9$\pm$3.0      & 96.7$\pm$2.3      & 97.3$\pm$1.8      & 96.9$\pm$1.7\\
        \midrule
        F-MNIST      		& 0     & 89.5$\pm$6.1      & 89.3$\pm$6.2      & 90.2$\pm$6.3  & 85.5$\pm$7.7      & 81.8$\pm$11.2 & 77.8$\pm$12.0     & 77.4$\pm$11.1 & 76.2$\pm$7.3      & 89.5$\pm$6.1      & 89.3$\pm$6.2      &                   & 76.2$\pm$7.3      & \\
        	 				& 1     &                   &                   &               &                   &               &                   &               &                   & 85.6$\pm$7.0      & 79.8$\pm$9.0      & 72.5$\pm$15.5     & 78.2$\pm$9.1      & 56.7$\pm$4.1\\
        					& 2     &                   &                   &               &                   &               &                   &               &                   & 87.8$\pm$6.1      & 80.1$\pm$10.5     & 74.3$\pm$15.4     & 80.5$\pm$8.2      & 62.3$\pm$2.9\\
        	 				& 3     &                   &                   &               &                   &               &                   &               &                   & 89.4$\pm$5.5      & 83.8$\pm$9.4      & 77.5$\pm$14.7     & 83.9$\pm$7.4      & 67.3$\pm$3.0\\
        	 				& 5     &                   &                   &               &                   &               &                   &               &                   & 91.2$\pm$4.8      & 86.8$\pm$7.7      & 79.9$\pm$13.8     & 87.3$\pm$6.4      & 75.3$\pm$2.7\\
        \midrule
        CIFAR-10	 		& 0     & 60.8$\pm$10.7     & 62.9$\pm$8.2      & 58.8$\pm$10.1 & 59.1$\pm$6.6      & 57.3$\pm$13.5 & 54.9$\pm$11.1     & 55.4$\pm$13.3 & 58.6$\pm$10.0     & 60.8$\pm$10.7     & 62.9$\pm$8.2      &                   & 58.6$\pm$10.0     & \\
        	 				& 1     &                   &                   &               &                   &               &                   &               &                   & 69.8$\pm$8.4      & 69.3$\pm$8.5      & 50.5$\pm$3.6      & 71.8$\pm$7.0      & 60.6$\pm$8.3\\
        					& 2     &                   &                   &               &                   &               &                   &               &                   & 73.0$\pm$7.1      & 72.3$\pm$7.5      & 50.3$\pm$2.4      & 75.2$\pm$6.4      & 61.0$\pm$6.6\\
        	 				& 3     &                   &                   &               &                   &               &                   &               &                   & 73.8$\pm$6.6      & 73.3$\pm$7.0      & 50.0$\pm$0.7      & 77.5$\pm$5.9      & 62.7$\pm$6.8\\
        	 				& 5     &                   &                   &               &                   &               &                   &               &                   & 75.1$\pm$5.5      & 74.2$\pm$6.5      & 50.0$\pm$1.0      & 80.4$\pm$4.6      & 60.9$\pm$4.6\\
        \bottomrule
        \end{tabular}
        %%%%%%%%%%%%%%%%%%%%%%%%%%%%%%%%%%%%%%%%
        \end{scriptsize}
    \end{landscape}
    \clearpage% Flush page
}
%%%%%%%%%%%%%%%%%%%%%%%%%%%%%%%%%%%%%%%%%%%%%%%%%%%%%%%%%%%%%%%%%%%%%%%%%%%%%%%%

%%%%%%%%%%%%%%%%%%%%%%%%%%%%%%%%%%%%%%%%%%%%%%%%%%%%%%%%%%%%%%%%%%%%%%%%%%%%%%%%
\afterpage{%
    \clearpage%
    \begin{landscape}% Landscape page
        \centering % Center table
        \captionsetup{type=table}
        \captionof{table}{Complete results on classic AD benchmark datasets in the setting with no pollution $\gamma_p = 0$ and a ratio of labeled anomalies of $\gamma_l = 0.01$ in the training set. We report the avg.~AUC with st.~dev.~computed over 10 seeds.}%
        \label{tab:odds_results_full}
        \vspace{1em}
        \begin{scriptsize}
        %%%%%%%%%%%%%%%%%%%%%%%%%%%%%%%%%%%%%%%%
        \begin{tabular}{lccccccccc}
        \toprule
                            & \textbf{OC-SVM}   & \textbf{OC-SVM}       &                       & \textbf{Deep}     & \textbf{SSAD}     & \textbf{SSAD}         &                   & \textbf{Deep}     & \textbf{Supervised} \\
        \textbf{Data}	    & \textbf{Raw}      & \textbf{Hybrid}       & \textbf{CAE}          & \textbf{SVDD}     & \textbf{Raw}      & \textbf{Hybrid}       & \textbf{SS-DGM}   & \textbf{SAD}      & \textbf{Classifier} \\
        \midrule
        arrhythmia          & 84.5$\pm$3.9      & 76.7$\pm$6.2          & 74.0$\pm$7.5          & 74.6$\pm$9.0      & 86.7$\pm$4.0      & 78.3$\pm$5.1          & 50.3$\pm$9.8      & 75.9$\pm$8.7      & 39.2$\pm$9.5 \\
        cardio              & 98.5$\pm$0.3      & 82.8$\pm$9.3          & 94.3$\pm$2.0          & 84.8$\pm$3.6      & 98.8$\pm$0.3      & 86.3$\pm$5.8          & 66.2$\pm$14.3     & 95.0$\pm$1.6      & 83.2$\pm$9.6 \\
        satellite           & 95.1$\pm$0.2      & 68.6$\pm$4.8          & 80.0$\pm$1.7          & 79.8$\pm$4.1      & 96.2$\pm$0.3      & 86.9$\pm$2.8          & 57.4$\pm$6.4      & 91.5$\pm$1.1      & 87.2$\pm$2.1 \\
        satimage-2          & 99.4$\pm$0.8      & 96.7$\pm$2.1          & 99.9$\pm$0.0          & 98.3$\pm$1.4      & 99.9$\pm$0.1      & 96.8$\pm$2.1          & 99.2$\pm$0.6      & 99.9$\pm$0.1      & 99.9$\pm$0.1 \\
        shuttle             & 99.4$\pm$0.9      & 94.1$\pm$9.5          & 98.2$\pm$1.2          & 86.3$\pm$7.5      & 99.6$\pm$0.5      & 97.7$\pm$1.0          & 97.9$\pm$0.3      & 98.4$\pm$0.9      & 95.1$\pm$8.0\\
        thyroid             & 98.3$\pm$0.9      & 91.2$\pm$4.0          & 75.2$\pm$10.2         & 72.0$\pm$9.7      & 97.9$\pm$1.9      & 95.3$\pm$3.1          & 72.7$\pm$12.0     & 98.6$\pm$0.9      & 97.8$\pm$2.6 \\
        \bottomrule
        \end{tabular}
        %%%%%%%%%%%%%%%%%%%%%%%%%%%%%%%%%%%%%%%%
        \end{scriptsize}
    \end{landscape}
    \clearpage% Flush page
}
%%%%%%%%%%%%%%%%%%%%%%%%%%%%%%%%%%%%%%%%%%%%%%%%%%%%%%%%%%%%%%%%%%%%%%%%%%%%%%%%

\end{document}

%% file: math_commands.tex
\usepackage{amsmath,amsfonts,bm}

\def\eqref#1{equation~\ref{#1}}
\def\1{\bm{1}}

\DeclareMathAlphabet{\mathsfit}{\encodingdefault}{\sfdefault}{m}{sl}
\SetMathAlphabet{\mathsfit}{bold}{\encodingdefault}{\sfdefault}{bx}{n}

%% file: main.bbl
\begin{thebibliography}{87}
\providecommand{\natexlab}[1]{#1}
\providecommand{\url}[1]{\texttt{#1}}
\expandafter\ifx\csname urlstyle\endcsname\relax
  \providecommand{\doi}[1]{doi: #1}\else
  \providecommand{\doi}{doi: \begingroup \urlstyle{rm}\Url}\fi

\bibitem[Achille \& Soatto(2018)Achille and Soatto]{achille2018}
Alessandro Achille and Stefano Soatto.
\newblock Emergence of invariance and disentanglement in deep representations.
\newblock \emph{Journal of Machine Learning Research}, 19\penalty0
  (1):\penalty0 1947--1980, 2018.

\bibitem[Akcay et~al.(2018)Akcay, Atapour-Abarghouei, and Breckon]{akcay2018}
Samet Akcay, Amir Atapour-Abarghouei, and Toby~P Breckon.
\newblock {GANomaly}: Semi-supervised anomaly detection via adversarial
  training.
\newblock In \emph{ACCV}, pp.\  622--637, 2018.

\bibitem[Alemi et~al.(2017)Alemi, Fischer, Dillon, and Murphy]{alemi2017}
Alexander~A Alemi, Ian Fischer, Joshua~V Dillon, and Kevin Murphy.
\newblock Deep variational information bottleneck.
\newblock In \emph{ICLR}, 2017.

\bibitem[Alemi et~al.(2018)Alemi, Poole, Fischer, Dillon, Saurous, and
  Murphy]{alemi2018}
Alexander~A Alemi, Ben Poole, Ian Fischer, Joshua~V Dillon, Rif~A Saurous, and
  Kevin Murphy.
\newblock Fixing a broken {ELBO}.
\newblock In \emph{ICML}, volume~80, pp.\  159--168, 2018.

\bibitem[Andrews et~al.(2016)Andrews, Morton, and Griffin]{andrews2016}
Jerone T~A Andrews, Edward~J Morton, and Lewis~D Griffin.
\newblock Detecting {A}nomalous {D}ata {U}sing {A}uto-{E}ncoders.
\newblock \emph{IJMLC}, 6\penalty0 (1):\penalty0 21, 2016.

\bibitem[Arora et~al.(2018)Arora, Cohen, and Hazan]{arora2018}
Sanjeev Arora, Nadav Cohen, and Elad Hazan.
\newblock On the optimization of deep networks: Implicit acceleration by
  overparameterization.
\newblock In \emph{ICML}, volume~80, pp.\  244--253, 2018.

\bibitem[Belkin et~al.(2018)Belkin, Ma, and Mandal]{belkin2018}
Mikhail Belkin, Siyuan Ma, and Soumik Mandal.
\newblock To understand deep learning we need to understand kernel learning.
\newblock In \emph{ICML}, pp.\  540--548, 2018.

\bibitem[Bell \& Sejnowski(1995)Bell and Sejnowski]{bell1995}
Anthony~J Bell and Terrence~J Sejnowski.
\newblock An information-maximization approach to blind separation and blind
  deconvolution.
\newblock \emph{Neural computation}, 7\penalty0 (6):\penalty0 1129--1159, 1995.

\bibitem[Blanchard et~al.(2010)Blanchard, Lee, and Scott]{blanchard2010}
Gilles Blanchard, Gyemin Lee, and Clayton Scott.
\newblock Semi-supervised novelty detection.
\newblock \emph{Journal of Machine Learning Research}, 11\penalty0
  (Nov):\penalty0 2973--3009, 2010.

\bibitem[Chalapathy \& Chawla(2019)Chalapathy and Chawla]{chalapathy2019}
Raghavendra Chalapathy and Sanjay Chawla.
\newblock Deep learning for anomaly detection: A survey.
\newblock \emph{arXiv:1901.03407}, 2019.

\bibitem[Chalapathy et~al.(2018)Chalapathy, Menon, and Chawla]{chalapathy2018}
Raghavendra Chalapathy, Aditya~Krishna Menon, and Sanjay Chawla.
\newblock Anomaly detection using one-class neural networks.
\newblock \emph{arXiv:1802.06360}, 2018.

\bibitem[Chandola et~al.(2009)Chandola, Banerjee, and Kumar]{chandola2009}
Varun Chandola, Arindam Banerjee, and Vipin Kumar.
\newblock Anomaly {D}etection: {A} {S}urvey.
\newblock \emph{ACM Computing Surveys}, 41\penalty0 (3):\penalty0 1--58, 2009.

\bibitem[Chapelle et~al.(2009)Chapelle, Sch{\"o}lkopf, and Zien]{chapelle2009}
Olivier Chapelle, Bernhard Sch{\"o}lkopf, and Alexander Zien.
\newblock Semi-supervised learning.
\newblock \emph{IEEE Transactions on Neural Networks}, 20\penalty0
  (3):\penalty0 542--542, 2009.

\bibitem[Chen et~al.(2017)Chen, Sathe, Aggarwal, and Turaga]{chen2017}
Jinghui Chen, Saket Sathe, Charu~C Aggarwal, and Deepak~S Turaga.
\newblock Outlier {D}etection with {A}utoencoder {E}nsembles.
\newblock In \emph{SDM}, pp.\  90--98, 2017.

\bibitem[Chen et~al.(2016)Chen, Duan, Houthooft, Schulman, Sutskever, and
  Abbeel]{chen2016}
Xi~Chen, Yan Duan, Rein Houthooft, John Schulman, Ilya Sutskever, and Pieter
  Abbeel.
\newblock {InfoGAN}: Interpretable representation learning by information
  maximizing generative adversarial nets.
\newblock In \emph{NIPS}, pp.\  2172--2180, 2016.

\bibitem[Cohen et~al.(2016)Cohen, Sharir, and Shashua]{cohen2016}
Nadav Cohen, Or~Sharir, and Amnon Shashua.
\newblock On the expressive power of deep learning: A tensor analysis.
\newblock In \emph{COLT}, volume~49, pp.\  698--728, 2016.

\bibitem[Cover \& Thomas(2012)Cover and Thomas]{cover2012}
Thomas~M Cover and Joy~A Thomas.
\newblock \emph{Elements of Information Theory}.
\newblock John Wiley \& Sons, 2012.

\bibitem[Dai et~al.(2017)Dai, Yang, Yang, Cohen, and Salakhutdinov]{dai2017}
Zihang Dai, Zhilin Yang, Fan Yang, William~W Cohen, and Ruslan~R Salakhutdinov.
\newblock Good semi-supervised learning that requires a bad gan.
\newblock In \emph{NIPS}, pp.\  6510--6520, 2017.

\bibitem[Deecke et~al.(2018)Deecke, Vandermeulen, Ruff, Mandt, and
  Kloft]{deecke2018}
Lucas Deecke, Robert~A Vandermeulen, Lukas Ruff, Stephan Mandt, and Marius
  Kloft.
\newblock Image anomaly detection with generative adversarial networks.
\newblock In \emph{ECML-PKDD}, 2018.

\bibitem[Denis(1998)]{denis1998}
Fran{\c{c}}ois Denis.
\newblock {PAC} learning from positive statistical queries.
\newblock In \emph{International Conference on Algorithmic Learning Theory},
  pp.\  112--126, 1998.

\bibitem[Eldan \& Shamir(2016)Eldan and Shamir]{eldan2016}
Ronen Eldan and Ohad Shamir.
\newblock The power of depth for feedforward neural networks.
\newblock In \emph{COLT}, volume~49, pp.\  907--940, 2016.

\bibitem[Erfani et~al.(2016)Erfani, Rajasegarar, Karunasekera, and
  Leckie]{erfani2016}
Sarah~M Erfani, Sutharshan Rajasegarar, Shanika Karunasekera, and Christopher
  Leckie.
\newblock High-dimensional and large-scale anomaly detection using a linear
  one-class {SVM} with deep learning.
\newblock \emph{Pattern Recognition}, 58:\penalty0 121--134, 2016.

\bibitem[Ergen et~al.(2017)Ergen, Mirza, and Kozat]{ergen2017}
Tolga Ergen, Ali~Hassan Mirza, and Suleyman~Serdar Kozat.
\newblock Unsupervised and semi-supervised anomaly detection with {LSTM} neural
  networks.
\newblock \emph{arXiv:1710.09207}, 2017.

\bibitem[Glorot \& Bengio(2010)Glorot and Bengio]{glorot2010}
Xavier Glorot and Yoshua Bengio.
\newblock Understanding the difficulty of training deep feedforward neural
  networks.
\newblock In \emph{AISTATS}, pp.\  249--256, 2010.

\bibitem[Golan \& El-Yaniv(2018)Golan and El-Yaniv]{golan2018}
Izhak Golan and Ran El-Yaniv.
\newblock Deep anomaly detection using geometric transformations.
\newblock In \emph{NeurIPS}, pp.\  9758--9769, 2018.

\bibitem[Goodfellow et~al.(2016)Goodfellow, Bengio, and
  Courville]{goodfellow2016}
Ian Goodfellow, Yoshua Bengio, and Aaron Courville.
\newblock \emph{Deep Learning}.
\newblock MIT Press, 2016.
\newblock \url{http://www.deeplearningbook.org}.

\bibitem[G{\"o}rnitz et~al.(2013)G{\"o}rnitz, Kloft, Rieck, and
  Brefeld]{gornitz2013}
Nico G{\"o}rnitz, Marius Kloft, Konrad Rieck, and Ulf Brefeld.
\newblock Toward supervised anomaly detection.
\newblock \emph{Journal of Artificial Intelligence Research}, 46:\penalty0
  235--262, 2013.

\bibitem[Hawkins et~al.(2002)Hawkins, He, Williams, and Baxter]{hawkins2002}
Simon Hawkins, Hongxing He, Graham Williams, and Rohan Baxter.
\newblock Outlier {D}etection {U}sing {R}eplicator {N}eural {N}etworks.
\newblock In \emph{DaWaK}, volume 2454, pp.\  170--180, 2002.

\bibitem[Hendrycks et~al.(2019{\natexlab{a}})Hendrycks, Mazeika, and
  Dietterich]{hendrycks2019}
Dan Hendrycks, Mantas Mazeika, and Thomas~G Dietterich.
\newblock Deep anomaly detection with outlier exposure.
\newblock In \emph{ICLR}, 2019{\natexlab{a}}.

\bibitem[Hendrycks et~al.(2019{\natexlab{b}})Hendrycks, Mazeika, Kadavath, and
  Song]{hendrycks2019using}
Dan Hendrycks, Mantas Mazeika, Saurav Kadavath, and Dawn Song.
\newblock Using self-supervised learning can improve model robustness and
  uncertainty.
\newblock In \emph{NeurIPS}, pp.\  15637--15648, 2019{\natexlab{b}}.

\bibitem[Hinton \& Salakhutdinov(2006)Hinton and Salakhutdinov]{hinton2006b}
Geoffrey~E Hinton and Ruslan~R Salakhutdinov.
\newblock Reducing the {D}imensionality of {D}ata with {N}eural {N}etworks.
\newblock \emph{Science}, 313\penalty0 (5786):\penalty0 504--507, 2006.

\bibitem[Hjelm et~al.(2019)Hjelm, Fedorov, Lavoie-Marchildon, Grewal,
  Trischler, and Bengio]{hjelm2018}
R~Devon Hjelm, Alex Fedorov, Samuel Lavoie-Marchildon, Karan Grewal, Adam
  Trischler, and Yoshua Bengio.
\newblock Learning deep representations by mutual information estimation and
  maximization.
\newblock In \emph{ICLR}, 2019.

\bibitem[Hoffman \& Johnson(2016)Hoffman and Johnson]{hoffman2016}
Matthew~D Hoffman and Matthew~J Johnson.
\newblock {ELBO} surgery: yet another way to carve up the variational evidence
  lower bound.
\newblock In \emph{NIPS Workshop in Advances in Approximate Bayesian
  Inference}, 2016.

\bibitem[Ioffe \& Szegedy(2015)Ioffe and Szegedy]{ioffe2015}
Sergey Ioffe and Christian Szegedy.
\newblock Batch {N}ormalization: {A}ccelerating {D}eep {N}etwork {T}raining by
  {R}educing {I}nternal {C}ovariate {S}hift.
\newblock In \emph{ICML}, pp.\  448--456, 2015.

\bibitem[Ji et~al.(2018)Ji, Henriques, and Vedaldi]{ji2018}
Xu~Ji, Joao~F Henriques, and Andrea Vedaldi.
\newblock Invariant information distillation for unsupervised image
  segmentation and clustering.
\newblock \emph{arXiv:1807.06653}, 2018.

\bibitem[Kim \& Scott(2012)Kim and Scott]{kim2012}
JooSeuk Kim and Clayton~D Scott.
\newblock Robust kernel density estimation.
\newblock \emph{Journal of Machine Learning Research}, 13\penalty0
  (Sep):\penalty0 2529--2565, 2012.

\bibitem[Kingma \& Ba(2015)Kingma and Ba]{kingma2014}
Diederik~P Kingma and Jimmy Ba.
\newblock Adam: {A} {M}ethod for {S}tochastic {O}ptimization.
\newblock In \emph{ICLR}, 2015.

\bibitem[Kingma \& Welling(2013)Kingma and Welling]{kingma2013}
Diederik~P Kingma and Max Welling.
\newblock Auto-encoding variational bayes.
\newblock In \emph{ICLR}, 2013.

\bibitem[Kingma et~al.(2014)Kingma, Mohamed, Rezende, and Welling]{kingma2014b}
Diederik~P Kingma, Shakir Mohamed, Danilo~Jimenez Rezende, and Max Welling.
\newblock Semi-supervised learning with deep generative models.
\newblock In \emph{NIPS}, pp.\  3581--3589, 2014.

\bibitem[Kiran et~al.(2018)Kiran, Thomas, and Parakkal]{kiran2018}
B~Kiran, Dilip Thomas, and Ranjith Parakkal.
\newblock An overview of deep learning based methods for unsupervised and
  semi-supervised anomaly detection in videos.
\newblock \emph{Journal of Imaging}, 4\penalty0 (2):\penalty0 36, 2018.

\bibitem[Lapuschkin et~al.(2019)Lapuschkin, W{\"a}ldchen, Binder, Montavon,
  Samek, and M{\"u}ller]{lapuschkin2019}
Sebastian Lapuschkin, Stephan W{\"a}ldchen, Alexander Binder, Gr{\'e}goire
  Montavon, Wojciech Samek, and Klaus-Robert M{\"u}ller.
\newblock Unmasking clever hans predictors and assessing what machines really
  learn.
\newblock \emph{Nature communications}, 10\penalty0 (1):\penalty0 1096, 2019.

\bibitem[Linsker(1988)]{linsker1988}
Ralph Linsker.
\newblock Self-organization in a perceptual network.
\newblock \emph{IEEE Computer}, 21\penalty0 (3):\penalty0 105--117, 1988.

\bibitem[Liu et~al.(2008)Liu, Ting, and Zhou]{liu2008}
Fei~T Liu, Kai~M Ting, and Zhi-Hua Zhou.
\newblock Isolation {F}orest.
\newblock In \emph{ICDM}, pp.\  413--422, 2008.

\bibitem[Liu \& Zheng(2006)Liu and Zheng]{liu2006}
Yi~Liu and Yuan~F Zheng.
\newblock Minimum enclosing and maximum excluding machine for pattern
  description and discrimination.
\newblock In \emph{ICPR}, pp.\  129--132, 2006.

\bibitem[Makhzani \& Frey(2014)Makhzani and Frey]{makhzani2013}
Alireza Makhzani and Brendan Frey.
\newblock K-sparse autoencoders.
\newblock In \emph{ICLR}, 2014.

\bibitem[Makhzani et~al.(2015)Makhzani, Shlens, Jaitly, Goodfellow, and
  Frey]{makhzani2015}
Alireza Makhzani, Jonathon Shlens, Navdeep Jaitly, Ian Goodfellow, and Brendan
  Frey.
\newblock Adversarial autoencoders.
\newblock \emph{arXiv:1511.05644}, 2015.

\bibitem[Min et~al.(2018)Min, Long, Liu, Cui, Cai, and Ma]{min2018}
Erxue Min, Jun Long, Qiang Liu, Jianjing Cui, Zhiping Cai, and Junbo Ma.
\newblock {SU-IDS}: A semi-supervised and unsupervised framework for network
  intrusion detection.
\newblock In \emph{International Conference on Cloud Computing and Security},
  pp.\  322--334, 2018.

\bibitem[Montavon et~al.(2011)Montavon, Braun, and M{\"u}ller]{montavon2011}
Gr{\'e}goire Montavon, Mikio~L Braun, and Klaus-Robert M{\"u}ller.
\newblock Kernel analysis of deep networks.
\newblock \emph{Journal of Machine Learning Research}, 12\penalty0
  (Sep):\penalty0 2563--2581, 2011.

\bibitem[Moya et~al.(1993)Moya, Koch, and Hostetler]{moya1993}
Mary~M Moya, Mark~W Koch, and Larry~D Hostetler.
\newblock One-class classifier networks for target recognition applications.
\newblock In \emph{Proceedings World Congress on Neural Networks}, pp.\
  797--801, 1993.

\bibitem[Mu{\~n}oz-Mar{\'\i} et~al.(2010)Mu{\~n}oz-Mar{\'\i}, Bovolo,
  G{\'o}mez-Chova, Bruzzone, and Camp-Valls]{munoz2010}
Jordi Mu{\~n}oz-Mar{\'\i}, Francesca Bovolo, Luis G{\'o}mez-Chova, Lorenzo
  Bruzzone, and Gustavo Camp-Valls.
\newblock Semi-{S}upervised {O}ne-{C}lass {S}upport {V}ector {M}achines for
  {C}lassification of {R}emote {S}ensing {S}ata.
\newblock \emph{IEEE Transactions on Geoscience and Remote Sensing},
  48\penalty0 (8):\penalty0 3188--3197, 2010.

\bibitem[Neyshabur et~al.(2017)Neyshabur, Bhojanapalli, McAllester, and
  Srebro]{neyshabur2017}
Behnam Neyshabur, Srinadh Bhojanapalli, David McAllester, and Nati Srebro.
\newblock Exploring generalization in deep learning.
\newblock In \emph{NIPS}, pp.\  5947--5956, 2017.

\bibitem[Nicolau et~al.(2016)Nicolau, McDermott, et~al.]{nicolau2016}
Miguel Nicolau, James McDermott, et~al.
\newblock A hybrid autoencoder and density estimation model for anomaly
  detection.
\newblock In \emph{International Conference on Parallel Problem Solving from
  Nature}, pp.\  717--726, 2016.

\bibitem[Odena(2016)]{odena2016}
Augustus Odena.
\newblock Semi-supervised learning with generative adversarial networks.
\newblock \emph{arXiv:1606.01583}, 2016.

\bibitem[Oliver et~al.(2018)Oliver, Odena, Raffel, Cubuk, and
  Goodfellow]{oliver2018}
Avital Oliver, Augustus Odena, Colin Raffel, Ekin~D Cubuk, and Ian~J
  Goodfellow.
\newblock Realistic evaluation of deep semi-supervised learning algorithms.
\newblock In \emph{NeurIPS}, pp.\  3235--3246, 2018.

\bibitem[Pang et~al.(2019)Pang, Shen, and van~den Hengel]{pang2019}
Guansong Pang, Chunhua Shen, and Anton van~den Hengel.
\newblock Deep anomaly detection with deviation networks.
\newblock In \emph{KDD}, pp.\  353--362, 2019.

\bibitem[Parzen(1962)]{parzen1962}
E.~Parzen.
\newblock On {E}stimation of a {P}robability {D}ensity {F}unction and {M}ode.
\newblock \emph{The annals of mathematical statistics}, 33\penalty0
  (3):\penalty0 1065--1076, 1962.

\bibitem[Pimentel et~al.(2014)Pimentel, Clifton, Clifton, and
  Tarassenko]{pimentel2014}
Marco~AF Pimentel, David~A Clifton, Lei Clifton, and Lionel Tarassenko.
\newblock A review of novelty detection.
\newblock \emph{Signal Processing}, 99:\penalty0 215--249, 2014.

\bibitem[Raghu et~al.(2017)Raghu, Poole, Kleinberg, Ganguli, and
  Dickstein]{raghu2017}
Maithra Raghu, Ben Poole, Jon Kleinberg, Surya Ganguli, and Jascha~Sohl
  Dickstein.
\newblock On the expressive power of deep neural networks.
\newblock In \emph{ICML}, volume~70, pp.\  2847--2854, 2017.

\bibitem[Rasmus et~al.(2015)Rasmus, Berglund, Honkala, Valpola, and
  Raiko]{rasmus2015}
Antti Rasmus, Mathias Berglund, Mikko Honkala, Harri Valpola, and Tapani Raiko.
\newblock Semi-supervised learning with ladder networks.
\newblock In \emph{NIPS}, pp.\  3546--3554, 2015.

\bibitem[Rayana(2016)]{rayana2016}
Shebuti Rayana.
\newblock {ODDS} library, 2016.
\newblock URL \url{http://odds.cs.stonybrook.edu}.

\bibitem[Rezende et~al.(2014)Rezende, Mohamed, and Wierstra]{rezende2014}
Danilo~Jimenez Rezende, Shakir Mohamed, and Daan Wierstra.
\newblock {Stochastic Backpropagation and Approximate Inference in Deep
  Generative Models}.
\newblock In \emph{ICML}, volume~32, pp.\  1278--1286, 2014.

\bibitem[Ruff et~al.(2018)Ruff, Vandermeulen, G{\"o}rnitz, Deecke, Siddiqui,
  Binder, M{\"u}ller, and Kloft]{ruff2018}
Lukas Ruff, Robert~A Vandermeulen, Nico G{\"o}rnitz, Lucas Deecke, Shoaib~A
  Siddiqui, Alexander Binder, Emmanuel M{\"u}ller, and Marius Kloft.
\newblock Deep one-class classification.
\newblock In \emph{ICML}, volume~80, pp.\  4390--4399, 2018.

\bibitem[Ruff et~al.(2019)Ruff, Zemlyanskiy, Vandermeulen, Schnake, and
  Kloft]{ruff2019}
Lukas Ruff, Yury Zemlyanskiy, Robert Vandermeulen, Thomas Schnake, and Marius
  Kloft.
\newblock Self-attentive, multi-context one-class classification for
  unsupervised anomaly detection on text.
\newblock In \emph{ACL}, pp.\  4061--4071, 2019.

\bibitem[Rumelhart et~al.(1986)Rumelhart, Hinton, and Williams]{rumelhart1986}
David~E Rumelhart, Geoffrey~E Hinton, and Ronald~J Williams.
\newblock Learning internal representations by error propagation.
\newblock In \emph{Parallel Distributed Processing -- Explorations in the
  Microstructure of Cognition}, chapter~8, pp.\  318--362. MIT Press, 1986.

\bibitem[Sakurada \& Yairi(2014)Sakurada and Yairi]{sakurada2014}
Mayu Sakurada and Takehisa Yairi.
\newblock Anomaly detection using autoencoders with nonlinear dimensionality
  reduction.
\newblock In \emph{Proceedings of the 2nd MLSDA Workshop}, pp.\ ~4, 2014.

\bibitem[Saxe et~al.(2018)Saxe, Bansal, Dapello, Advani, Kolchinsky, Tracey,
  and Cox]{saxe2018}
Andrew~Michael Saxe, Yamini Bansal, Joel Dapello, Madhu Advani, Artemy
  Kolchinsky, Brendan~Daniel Tracey, and David~Daniel Cox.
\newblock On the information bottleneck theory of deep learning.
\newblock In \emph{ICLR}, 2018.

\bibitem[Sch{\"o}lkopf \& Smola(2002)Sch{\"o}lkopf and
  Smola]{scholkopf2001learning}
Bernhard Sch{\"o}lkopf and Alex~J Smola.
\newblock \emph{Learning with Kernels}.
\newblock MIT press, 2002.

\bibitem[Sch{\"o}lkopf et~al.(2001)Sch{\"o}lkopf, Platt, Shawe-Taylor, Smola,
  and Williamson]{scholkopf2001}
Bernhard Sch{\"o}lkopf, John~C Platt, John Shawe-Taylor, Alex~J Smola, and
  Robert~C Williamson.
\newblock Estimating the {S}upport of a {H}igh-{D}imensional {D}istribution.
\newblock \emph{Neural computation}, 13\penalty0 (7):\penalty0 1443--1471,
  2001.

\bibitem[Scott \& Nowak(2006)Scott and Nowak]{scott2006}
Clayton~D Scott and Robert~D Nowak.
\newblock Learning minimum volume sets.
\newblock \emph{Journal of Machine Learning Research}, 7\penalty0
  (Apr):\penalty0 665--704, 2006.

\bibitem[Shannon(1948)]{shannon1948}
Claude~Elwood Shannon.
\newblock A mathematical theory of communication.
\newblock \emph{Bell system technical journal}, 27\penalty0 (3):\penalty0
  379--423, 1948.

\bibitem[Shwartz-Ziv \& Tishby(2017)Shwartz-Ziv and Tishby]{shwartz2017}
Ravid Shwartz-Ziv and Naftali Tishby.
\newblock Opening the black box of deep neural networks via information.
\newblock \emph{arXiv:1703.00810}, 2017.

\bibitem[Slonim et~al.(2005)Slonim, Atwal, Tka{\v{c}}ik, and
  Bialek]{slonim2005}
Noam Slonim, Gurinder~Singh Atwal, Ga{\v{s}}per Tka{\v{c}}ik, and William
  Bialek.
\newblock Information-based clustering.
\newblock \emph{Proceedings of the National Academy of Sciences}, 102\penalty0
  (51):\penalty0 18297--18302, 2005.

\bibitem[Song et~al.(2017)Song, Jiang, Men, and Yang]{song2017}
Hongchao Song, Zhuqing Jiang, Aidong Men, and Bo~Yang.
\newblock A hybrid semi-supervised anomaly detection model for high-dimensional
  data.
\newblock \emph{Computational Intelligence and Neuroscience}, 2017.

\bibitem[Steinwart et~al.(2005)Steinwart, Hush, and Scovel]{steinwart2005}
Ingo Steinwart, Don Hush, and Clint Scovel.
\newblock A classification framework for anomaly detection.
\newblock \emph{Journal of Machine Learning Research}, 6\penalty0
  (Feb):\penalty0 211--232, 2005.

\bibitem[Tax \& Duin(2004)Tax and Duin]{tax2004}
David M~J Tax and Robert P~W Duin.
\newblock Support {V}ector {D}ata {D}escription.
\newblock \emph{Machine learning}, 54\penalty0 (1):\penalty0 45--66, 2004.

\bibitem[Tishby \& Zaslavsky(2015)Tishby and Zaslavsky]{tishby2015}
Naftali Tishby and Noga Zaslavsky.
\newblock Deep learning and the information bottleneck principle.
\newblock In \emph{2015 IEEE Information Theory Workshop (ITW)}, pp.\  1--5,
  2015.

\bibitem[Tishby et~al.(1999)Tishby, Pereira, and Bialek]{tishby2000}
Naftali Tishby, Fernando~C Pereira, and William Bialek.
\newblock The information bottleneck method.
\newblock In \emph{The 37th annual Allerton Conference on Communication,
  Control and Computing}, pp.\  368--377, 1999.

\bibitem[Vandermeulen \& Scott(2013)Vandermeulen and Scott]{vandermeulen2013}
Robert~A Vandermeulen and Clayton Scott.
\newblock Consistency of robust kernel density estimators.
\newblock In \emph{COLT}, pp.\  568--591, 2013.

\bibitem[Vincent et~al.(2008)Vincent, Larochelle, Bengio, and
  Manzagol]{vincent2008}
Pascal Vincent, Hugo Larochelle, Yoshua Bengio, and Pierre-Antoine Manzagol.
\newblock Extracting and composing robust features with denoising autoencoders.
\newblock In \emph{ICML}, pp.\  1096--1103, 2008.

\bibitem[Wang et~al.(2005)Wang, Neskovic, and Cooper]{wang2005}
Jigang Wang, Predrag Neskovic, and Leon~N Cooper.
\newblock Pattern classification via single spheres.
\newblock In \emph{International Conference on Discovery Science}, pp.\
  241--252. Springer, 2005.

\bibitem[Wiatowski \& B{\"o}lcskei(2018)Wiatowski and
  B{\"o}lcskei]{wiatowski2018}
Thomas Wiatowski and Helmut B{\"o}lcskei.
\newblock A mathematical theory of deep convolutional neural networks for
  feature extraction.
\newblock \emph{IEEE Transactions on Information Theory}, 64\penalty0
  (3):\penalty0 1845--1866, 2018.

\bibitem[Wilcoxon(1945)]{wilcoxon1945}
Frank Wilcoxon.
\newblock Individual comparisons by ranking methods.
\newblock \emph{Biometrics Bulletin}, 1\penalty0 (6):\penalty0 80--83, 1945.

\bibitem[Zhai et~al.(2016)Zhai, Cheng, Lu, and Zhang]{zhai2016}
Shuangfei Zhai, Yu~Cheng, Weining Lu, and Zhongfei Zhang.
\newblock Deep structured energy based models for anomaly detection.
\newblock In \emph{ICML}, volume~48, pp.\  1100--1109, 2016.

\bibitem[Zhang \& Zuo(2008)Zhang and Zuo]{zhang2008}
Bangzuo Zhang and Wanli Zuo.
\newblock Learning from positive and unlabeled examples: A survey.
\newblock In \emph{Proceedings of the {IEEE} International Symposium on
  Information Processing}, pp.\  650--654, 2008.

\bibitem[Zhang et~al.(2017)Zhang, Bengio, Hardt, Recht, and Vinyals]{zhang2016}
Chiyuan Zhang, Samy Bengio, Moritz Hardt, Benjamin Recht, and Oriol Vinyals.
\newblock Understanding deep learning requires rethinking generalization.
\newblock In \emph{ICLR}, 2017.

\bibitem[Zhao et~al.(2017)Zhao, Song, and Ermon]{zhao2017}
Shengjia Zhao, Jiaming Song, and Stefano Ermon.
\newblock {InfoVAE}: Information maximizing variational autoencoders.
\newblock \emph{arXiv:1706.02262}, 2017.

\bibitem[Zhu(2005)]{zhu2005}
Xiaojin Zhu.
\newblock Semi-supervised learning literature survey.
\newblock \emph{Computer Sciences TR 1530, University of Wisconsin Madison},
  2005.

\end{thebibliography}
